\documentclass{sig-alternate}
\usepackage{times}
\usepackage{paralist}
\usepackage{balance}  % to better equalize the last page
\usepackage{graphicx} % for EPS, load graphicx instead
\usepackage{url}      % llt: nicely formatted URLs
\usepackage{amsmath}
\usepackage{color}
\usepackage{listings}
\usepackage{graphicx}
\usepackage[english]{babel}
\usepackage{booktabs}
\usepackage{url}      % llt: nicely formatted URLs
\usepackage{graphicx}
\usepackage[scriptsize, it, IT]{subfigure}
\usepackage[font={scriptsize,it}]{caption}
\usepackage{tikz}
\usetikzlibrary{arrows,matrix,positioning}
\usepackage[ruled,vlined,algonl,boxed]{algorithm2e}
\usepackage{framed}

\newcommand{\agp}[1]{\noindent{\textcolor{red}{Aditya: #1}}}

\newcommand{\stitle}[1]{\vspace{0.5em}\noindent\textbf{#1}}
\newcommand{\calF}[0]{$\cal{F}$}

\makeatletter
\def\url@leostyle{%
  \@ifundefined{selectfont}{\def\UrlFont{\sf}}{\def\UrlFont{\small\bf\ttfamily}}}
\makeatother
\def\pprw{8.5in}
\def\pprh{11in}

\newtheorem{definition}{Definition}[section]

\newtheorem{lemma}[definition]{Lemma}

\newtheorem{theorem}[definition]{Theorem}

\newcounter{prob}
\newtheorem{problem}[prob]{Problem}

\newtheorem{property}[definition]{Property}
\newtheorem{axiom}[definition]{Axiom}

\tikzset{
    >=stealth',
    punkt/.style={
           rectangle,
           rounded corners,
           draw=black, very thick,
           text width=6.5em,
           minimum height=2em,
           text centered},
    pil/.style={
           ->,
           thick,
           shorten <=2pt,
           shorten >=2pt,}
}

\newcommand{\squishlist}{
   \begin{list}{$\bullet$}
    { \setlength{\itemsep}{0pt}
      \setlength{\parsep}{2pt}
      \setlength{\topsep}{6pt}
      \setlength{\partopsep}{0pt}
      \leftmargin=25pt
\rightmargin=0pt
\labelsep=5pt
\labelwidth=10pt
\itemindent=0pt
\listparindent=0pt
\itemsep=\parsep
    }
}
\newcommand{\squishend}{\end{list}}

\newcommand{\squishframe}{\vspace{-6pt}
\begin{framed} 
\vspace{-6pt}}

\newcommand{\frameend}{\vspace{-6pt}
\end{framed}
\vspace{-6pt}}

\newcommand{\pdom}{{\sc Poly-Dom}\xspace}
\newcommand{\pdomind}{{\sc Poly-Dom-Index-All}\xspace}
\newcommand{\nlook}{{\sc Naive-Lookup}\xspace}

\newcommand{\nexp}{{\sc Naive-Expand-All}\xspace}
\newcommand{\gm}{{\sc Greedy}\xspace}
\newcommand{\gacc}{{\sc Greedy-Acc}\xspace}
\newcommand{\gcost}{{\sc Greedy-Cost}\xspace}

\usepackage[pdftex]{hyperref}
\hypersetup{
pdftitle={SIGCHI Conference Proceedings Format},
pdfauthor={LaTeX},
pdfkeywords={SIGCHI, proceedings, archival format},
bookmarksnumbered,
pdfstartview={FitH},
colorlinks,
citecolor=black,
filecolor=black,
linkcolor=black,
urlcolor=black,
breaklinks=true,
}

\begin{document}

% for 
\title{Indexing Cost Sensitive Prediction}

% \numberofauthors{5}
% \author{
%   \alignauthor Leilani Battle\\
%     \affaddr{MIT}\\
%     \affaddr{Address}\\
%   %  \email{e-mail address}
%   \alignauthor Ted Benson\\
%     \affaddr{MIT}\\
%     \affaddr{Address}\\
%   %  \email{e-mail address}
%   \alignauthor Eugene Wu\\
%     \affaddr{MIT}\\
%     \affaddr{Address}\\
%   %  \email{e-mail address}
%    \alignauthor Aditya Parameswaran\\
%     \affaddr{U. Illinois and MIT}\\
%     \affaddr{Address}\\
%   %  \email{e-mail address}
% }

% \numberofauthors{3}
% \author{
% \alignauthor \hspace{-40pt} Leilani Battle\\
% \affaddr{\hspace{-40pt}MIT}\\
% \affaddr{\hspace{-40pt}leibatt@mit.edu}
% \alignauthor \hspace{-80pt} Ted Benson\\
% \affaddr{\hspace{-80pt}MIT}\\
% \affaddr{\hspace{-80pt}eob@csail.mit.edu}
% \alignauthor \hspace{-105pt} Eugene Wu\\
% \affaddr{\hspace{-105pt}MIT}\\
% \affaddr{\hspace{-105pt}eugenewu@mit.edu}
% \alignauthor \hspace{-135pt} Aditya Parameswaran\\
% \affaddr{\hspace{-130pt}U. Illinois and MIT}\\
% \affaddr{\hspace{-130pt}adityagp@illinois.edu}
% }

% Teaser figure can go here
%\teaser{
%  \centering
%  \includegraphics{Figure1}
%  \caption{Teaser Image}
%  \label{fig:teaser}
%}

\maketitle
\vspace{-5in}

\begin{abstract} 

Predictive models are often used for real-time
decision making. However, typical machine learning techniques ignore
feature evaluation cost, and focus solely on the accuracy
of the machine learning models obtained utilizing all the features available. 
We develop algorithms and indexes to support
cost-sensitive prediction, i.e., making decisions 
using machine learning models taking feature evaluation
cost into account.
Given an item and a online computation cost (i.e., time) budget, we present two approaches
to return an appropriately chosen machine learning model
that will run within the specified time on the given item.
The first approach returns the optimal machine learning
model, i.e., one with the highest accuracy, that runs within the specified time,
but requires significant up-front precomputation time.
The second approach returns a possibly sub-optimal machine learning model,
but requires little up-front precomputation time.
We study these two algorithms in detail and characterize the scenarios 
(using real and synthetic data) in which each performs well.
Unlike prior work that focuses on a narrow domain or a specific algorithm, our 
techniques are {\em very general}: they apply to any cost-sensitive prediction scenario
on any machine learning algorithm. 
% In experiments, our techniques yielded a 
% reduction of XXX in cost while retaining 
% the same accuracy over other conventional schemes.
\end{abstract}

%!TEX root=main.tex

%!TEX root = ../main.tex

\section{Introduction}

Predictive models are ubiquitous in real-world applications:
ad-networks predict which ad the user will most likely click on
based on the user's web history, Netflix uses a user's viewing and
voting history to pick movies to recommend, and content moderation
services decide if an uploaded image is appropriate for young
children.  In these applications, the predictive model needs to
process the input data and make a prediction within a bounded amount
of time, or risk losing user engagement or revenue~\cite{brutlag,hamilton,shurman}.

Unfortunately, traditional feature-based classifiers take a one-model-fits all approach when placed in production, behaving the same way regardless of input size or time budget.
From the classifier's perspective, the features used to represent an input item have already been computed, and this computation process is external to the core task of classification.
This approach isolates and simplifies the core task of machine learning, allowing theorists to focus on tasks like quick learning convergence and accuracy.
But it leaves out many important aspects of production systems can be just as important as accuracy, such as tunable prediction speed.

In reality, the cost of computing features can easily dominate prediction time.
For example, a content moderation application may use a support-vector machine to detect inappropariate images. 
At runtime, SVMs need only perform a single dot-product between a feature vector and pre-computed weight vector.
But computing the feature vector may require several scans of an image which may take longer than an alotted time budget.

If feature computation is the dominating cost factor, one might intuitively accomodate time constraints by computing and using only a subset of features available.
But selecting which subset to use at runtime---and guaranteeing that a model  is available that was trained on that subet---is made challenging by a number of factors.
Features vary in \textit{predictive power}, e.g. skin tone colors might more accurately predict inappropriate images than image size.
They also vary in \textit{prediction cost}, e.g. looking up the image
size is much faster than computing a color histogram over an entire
image.
This cost also varies with respect to input size---the size feature may be $O(1)$, stored in metadata, while the histogram may be $O(r \times c)$.
Finally for any $n$ features, $2^n$ distinct subsets are possible, each with their aggregate predictive power and cost, and each potentially requiring its own custom training run.
As the number of potential features grows large, training a model for every possible subset is clearly prohibitively costly. 
All of these reasons highlight why deploying real-time prediction, while extremely important, is a particularly challenging problem.

Existing strategies of approaching this problem (see Section~\ref{sec:related}) 
tend to be either tightly coupled to a particular prediction task or to a particular mathematical model.
% In either case, the cost-sensitive prediction adaptation relies on this knowledge to construct a system that lends well to a greedy approach, in which a cascade of decisions are made, each refining the decision of the previous one.
While these approaches work for a particular problem, they are narrow in their applicability: if the domain (features) change or the machine learning model is swapped for new one (e.g., SVM for AdaBoost), the approach will no longer work.

In this paper, we develop a framework for {\em cost-sensitive real-time classification} 
as a {\em wrapper} over ``off-the-shelf'' feature-based classifiers.
That is, given an item that needs to be classified or categorized 
in real time and a cost (i.e., time) budget for feature evaluation, our goal is
to {\em identify features to compute in real time that are within the budget, 
identify the appropriate machine learning model that has been learned in advance, 
and apply the model on the extracted features.}

We take a systems approach by decoupling the problem of cost-sensitive prediction from the problem of machine learning in general.
We present an algorithm for cost sensitive prediction that operates on any feature-based machine learning algorithm as a black box.
The few assumptions it makes reasonably transfer between different feature sets and algorithms (and are justified herein).
This decoupled approach is attractive for the same reason that machine learning literature did not originally address such problems: it segments reasoning about the core tasks of learning and prediction from system concerns about operationalizing and scaling.
Additionally,  encapsulating the details of classification as we do ensures advances in machine learning algorithms and feature engineering can be integrated without change to the cost-sensitivity apparatus.

Thus, our focus in this paper is on {\em systems issues underlying this wrapper-based approach}, i.e., 
on intelligent indexing and pruning techniques to enable
rapid online decision making and not on the machine learning algorithms themselves.
Our contribution is two approaches to this problem as well as new techniques to mitigate the challenges of each approach.
These two approaches represents two ends of a continuum of approaches to tackle the problem of model-agnostic cost sensitivity:

\begin{compactitem}
\item Our \pdom approach yields optimal
solutions but requires significant offline pre-computation, 

% is an index that stores, for each input size, a skyline of predictive models along the axes of total real-time computation cost versus accuracy. 
% At test time, given an input, we can simply pick the best predictive model along the skyline within the total computation cost budget.
% We show that the \pdom approach yields optimal solutions but requires significant offline pre-computation.

\item Our \gm approach yields
relatively good solutions but does not require
significant offline pre-computation.

 % (adapted from prior work) good, but often suboptimal, solutions but does not require the significant pre-computation of the \pdom.
\end{compactitem}
First, consider the \gm approach:
\gm, and its two sub-variants \gacc and \gcost
(described in Section~\ref{sec:gm-solution}),
are all simple but effective techniques
adapted from prior work by Xu {\em et al.}~\cite{DBLP:conf/icml/XuWC12},
wherein the technique only applied to a sub-class of SVMs~\cite{hearst1998support}.
Here, we generalize the techniques to apply to
any machine learning classiciation algorithm as a black box.
\gm is a ``quick and dirty'' technique that requires little precomputation,
storage and retrieval, and works well in many settings.

Then, consider our \pdom approach,
which is necessary whenever accuracy is paramount,
a typical scenario in critical applications like
credit card fraud detection, system performance monitoring, 
and ad-serving systems.
In this approach, we conceptually store, 
for each input size, a skyline of predictive
models along the axes of total real-time computation cost vs. accuracy.
Then, given an input of that size, we can simply pick the predictive model
along the skyline
within the total real-time computation cost budget, 
and then get the best possible
accuracy.

% \agp{Alternative 1:}

However, there are many difficulties in implementing this skyline-based approach:
\begin{compactitem}
\item Computing the skyline in a naive fashion requires us to 
compute for all $2^{|\cal{F}|}$,  
subsets of features (where $\cal{F}$ is the set of all features) 
the best machine learning
algorithm for that set, and the total real-time computation time (or cost).
If the number of features is large, say in the 100s or the 1000s, computing
the skyline is impossible to do, even with an unlimited amount of time offline. 
How do we intelligently reduce the amount of precomputation required to find the skyline?

\item The skyline, once computed, will require a lot of storage. How 
should this skyline be stored, and what index structures
should we use to allow efficient retrieval of individual models on the skyline?

\item Computing and storing the skyline for each input size is simply infeasible:
an image, for instance, can vary between 0 to 70 Billion Pixels (the size of the largest 
photo on earth~\cite{largest}), we simply cannot store or precompute
this much information. What can we do in such a case?

\end{compactitem}
To deal with the challenges above, \pdom use a dual-pronged solution, with two 
precomputation steps:
\begin{compactitem}
\item {\em Feature Set Pruning:} We develop a number of pruning techniques 
that enable us to minimize the number of feature sets for which we need
to learn machine learning algorithms. Our lattice pruning techniques
are {\em provably correct under some very reasonable assumptions}, i.e.,
they do not discard any feature sets if those feature sets could be
potentially optimal under certain input conditions. 
We find that our pruning techniques often allow us to prune up to 90\% of
the feature sets.
\item {\em Polydom Index:} Once we gather the collection of feature sets,
we develop an index structure that allows us to represent the models
learned using the feature sets in such a way that enables us to perform
{\em efficient retrieval of the optimal machine learning model given
constraints on cost and input size}. 
This index structure relies on reasoning about polynomials that represent
cost characteristics of feature sets as a function of input size.
\end{compactitem}

% \agp{Alternative 2:}
% We additionally contribute techniques for mitigating the challenges of building and maintaining the data structures necessary to support the higher-performing \pdom approach.
% We develop a number of lattice-based pruning techniques that enable us to minimize the number of models we need to train and store.
% These techniques are shown to be provably correct under some reasonable assumptions, i.e., they do not discard any feature sets if those feature sets could be potentially optimal under certain input conditions.
% We find that these techniques allow us to prune up to XXX\%\agp{yo yo fix} of the feature sets.

% Once we have computed a lattice of models, we develop an index structure that permits efficient storage and retrieval of models at run-time.
% This index structure relies on reasoning about polynomials that represent
% cost characteristics of feature sets as a function of input size.
% This allows fast one-shot execution within time budgets, as opposed to cascading computational models pursued by other approaches to this problem.

% \agp{End Alternatives}

Overall, our approach offers a systems perspective to an increasingly important topic in the deployment of machine learning systems.
The higher-level goal is to isolate and develop the mechanisms for storage and delivery of cost-sensitive prediction without having to break the encapsulation barrier that should surround the fundamental machinery of machine learning. 
Our techniques could be deployed alongside the existing algorithms in a variety of real-time prediction scenarios, including:

\begin{compactitem}

\item An ad system needs to balance between per-user ad customization and latency on the small scale, and allocate computational resources between low-value and high-value ad viewers on the aggregate scale.

%  to quickly serve users the most
% profitable ads. However user profiles and histories vary significantly
% between users and ads, and the computational resources for computing
% features from user and ad data varies depending on the current
% system load (e.g., middle of the night vs peak hours).

\item Cloud-based financial software needs to run predictive models on portfolios of dramatically different input size and value. A maximum latency on results may be required, but the best possible model for each time and input size pairing is financially advantageous. 

 % financial advice  a recommendation system to provide financial advice when users upload their financial histories.  The system needs to quickly generate results before users get bored.

\item An autopilot system in an airplane has a limited time to
respond to an error. Or more broadly, system performance monitors in a variety of industrial systems have fixed time to decide whether to alert a human operator of a failure.

\item A mobile sensor has limited resources to decide if an error
needs to be flagged and sent to the central controller.

% \item A system performance monitor has a fixed time to decide whether
% to alert an administrator during a system failure.

% \item A content moderation system on Facebook or Flickr has a limited
% time to decide if a user-uploaded image should be displayed on Facebook or Flickr.

% ewu: too soon
%\item A missile defense system has a fixed time to calculate a
%firing decision.

%\item A credit card transaction system has limited time
%to verify if a transaction is fraudulent or not.

\end{compactitem}

In the rest of the paper, we will first formally present our problem
variants in Section~\ref{sec:setup}, then describe our two-pronged \pdom
solution in Section~\ref{sec:sol} and our ``quick and dirty'' \gm solution
in Section~\ref{sec:gm-solution}, and finally present our experiments
on both synthetic and real-world datasets in Section~\ref{sec:exp}.

%!TEX root=../main.tex

\section{Problem Description}\label{sec:setup}

We begin by describing some notation that will apply to the rest of the
paper, and then we will present the formal statement of the problems
that we study.

Our goal is to classify an item (e.g., image, video, text) $x$ during real-time.
We assume that the size of $x$, denoted $|x|$, would be represented using a 
single number or dimension $n$, e.g., number of words in the text. 
Our techniques also apply to the scenario when the size can be 
represented using a vector of dimensions: for example, (length, breadth), for
an image; however, for ease of exposition, we focus on 
the single dimension scenario. 
The entire set of features we can evaluate on $x$ is ${\cal F}$;
each individual feature is denoted $f_i$, while a non-empty set of features
is denoted $F_j$.

We assume that we have some training data, denoted ${\cal T}$, wherein
every single feature $f_i \in {\cal F}$ is evaluated for each 
item. 
Since training is done offline, it is not unreasonable to expect that
we have the ability to compute all the features on each item.
In addition, we have some testing data, denoted ${\cal T'}$, where once
again every single feature is evaluated for each item.
We use this test data to estimate the accuracy of the machine
learning models we discover offline.

\stitle{Cost Function:}
We assume that evaluating a feature on an item $x$ depends only on the
feature that is being computed, and the size of the item $|x|$.
We denote the cost of computing $f_i$ on $x$ as: $c(f_i, |x|)$.
We can estimate $c(f_i, n)$ during pre-processing time
by running the subroutine corresponding to feature evaluation $f_i$
on varying input sizes. 
Our resulting expression for $c(f_i, n)$ could either
be a constant (if it takes a fixed amount of time to 
evaluate the feature, no matter the size),
or could be a function of $n$, e.g., $3n^2 + 50$,
if evaluating a feature depends on $n$.

Then, the cost of computing a set of features $F_i$ on $x$ can be computed 
as follows:
\begin{equation}\label{eq:cost-model}
c(F_i, |x|) = \sum_{f \in F_i} c(f, |x|)
\end{equation}
We assume that each feature is computed independently, in sequential order.
Although there may be cases where multiple features can be computed together
(e.g., multiple features can share scans over an image simultaneously),
we expect that users provide the features as ``black-box'' subroutines and do not 
want to place additional burden
by asking users to provide subroutines for combinations of features as well.
That said, our techniques will equally well apply to the scenario when 
our cost model is more general than Equation~\ref{eq:cost-model}, or 
if users have provided subroutines for generating multiple
feature values simultaneously (e.g., extracting a word frequency vector from a text document).

\stitle{Accuracy Function:}
We model the machine learning algorithm (e.g., SVM, decision tree, naive-bayes) 
as a black box function supplied by the user.
This algorithm takes as input the entire training data ${\cal T}$, 
as well as a set of features $F_i$,
and outputs the best model ${\cal M}(F_i)$ 
learned using the set of features $F_i$. 
We denote the accuracy of ${\cal M}(F_i)$ inferred on the testing data
${\cal T'}$ as $a(F_i)$, possibly 
using k-fold cross-validation. 
We assume that the training data is representative of the items classified 
online (as is typical), so that the accuracy of the model ${\cal M}(F_i)$ is still $a(F_i)$ 
online.

%For any item classified online using the model
%${\cal M}(F_i)$, we assume that (as is typical) that the probability
%of correct classification is still $a(F_i)$.

Note that we are implicitly assuming that the accuracy of the
classification model only depends on the set of features inferred during test time,
and not on the size of the item.
% \ewu{Many features such as SIFT features are independent of the data item size.  However others
% susch as word vectors are not...}
This assumption is typically true in practice: whether
or not an image needs to be flagged for moderation is
independent of the size of the image.
% \agp{Any better reason why this is a reasonable assumption?}
% \ewu{is determining accuracy(item size) an open problem?}
% \ewu{could extend our model to learn accuracy functions with minor changes in
% skyline algorithms?}

\stitle{Characterizing a Feature Set:} 
Since we will be dealing often with sets of features at a time, 
we now describe what we mean by characterizing 
a feature set $F_i$. Overall, given $F_i$, 
as discussed above, we have a black box that
returns 
\begin{compactitem}
\item a machine learning model learned on some or all the features in $F_i$, 
represented as ${\cal M}(F_i)$. 
\item $a(F_i)$, i.e., an estimate of the accuracy of the model ${\cal M} (F_i)$ on test data ${\cal T'}$.
\end{compactitem}
In addition, we can estimate 
$c(F_i, n)$, i.e., the cost of extracting the features
to apply the 
model at test time as a function of the size
of the item $n$. 
Note that unlike the last two quantities, this quantity
will be expressed in symbolic form.
For example, $c(F_i, n) $ could be an expression like $3 n^2 + 4 n \log n + 7 n$.

Characterizing a feature set $F_i$ thus involves learning all three quantities
above for $F_i$: ${\cal M}, a, c$.
For the rest of the paper, we will operate on feature sets,
implicitly assuming that a feature set is characterized by
the best machine learning model for that feature set, an accuracy
value for that model, and a cost function.

\stitle{Problem Statements:}
The most general version of the problem is when $c$ (i.e., the cost or time constraint) 
and the size $n$ of an item $x$ are not provided to us in advance:
\begin{problem}[Prob-General]\label{prob}
Given ${\cal F}, {\cal T, T'}$ at preprocessing time, compute classification models
and indexes such that the following task can be completed at real-time:
\begin{compactitem}
\item Given $x, |x| = n$, and a cost constraint $c$ at real time, identify a set 
$F_i \in {\cal F'}=\{F \subseteq {\cal F}\ |\ c(F, n) \le c\}$ such that
$\forall_{F \in {\cal F'}} \alpha\times a(F_i) \ge a(F)$ and return ${\cal M}(F_i)(x)$.

%$c(F_i, n_0) \leq c_0$, and from among all $F_j$ such that $c(F_j, n_0) \leq c_0$,
%$\alpha a(F_i) \geq a(F_j)$, and return ${\cal M} (F_i) (x)$.
\end{compactitem}
\end{problem}
That is, our goal is to build classification models and indexes such that given
a new item at real time, we select a feature set $F_i$
and the corresponding machine learning model ${\cal M} (F_i)$
that both obeys the cost constraint, 
and is within $\alpha \geq 1\times$ of the best accuracy among all feature sets
that obey the cost constraint.
The reason we care about $\alpha >1$ is
that, in contrast to a hard cost constraint, a slightly lower accuracy is often acceptable as 
long as the amount of computation required for computing, storing,
and retrieving the appropriate models is manageable.
We will consider $\alpha = 1$, i.e., the absolute best accuracy,
as a special case; however, for most of the paper,
we will consider the more general variants.

There are two special cases of the general problem that we will consider.
The first special case considers the scenario when the input
size is provided up-front, e.g., when
Yelp fixes the size of profile images uploaded 
to the website that need to be moderated.
\begin{problem}[Prob-$n_0$-Fixed]\label{prob-n}
Given ${\cal F}, {\cal T, T'}$ and a fixed $n_0$ at preprocessing time, compute classification models
and indexes such that the following task can be completed at real-time:
\begin{compactitem}
\item Given $x, |x| = n_0$, and a cost constraint $c$ at real time, identify the set 
$F_i \in {\cal F'}=\{F \subseteq {\cal F}\ |\ c(F, n_0) \le c\}$ such that
$\forall_{F \in {\cal F'}} \alpha\times a(F_i) \ge a(F)$ and return ${\cal M}(F_i)(x)$.
%$F_i \subseteq {\cal F}$, such that $c(F_i, n_0) \leq c_x$, 
%and from among all $F_j$ such that $c(F_j, n_0) \leq c_x$,
%$\alpha a(F_i) \geq a(F_j)$, and return ${\cal M} (F_i) (x)$.
\end{compactitem}
\end{problem}
We also consider the version where $c_0$ is provided in advance
but the item size $n$ is not, e.g., when an aircraft needs to respond to any signals
within a fixed time.
\begin{problem}[Prob-$c_0$-Fixed]\label{prob-c0}
Given $c_0, {\cal F}, {\cal T, T'}$ at preprocessing time, compute classification models
and indexes such that the following task can be completed at real-time:
\begin{compactitem}
\item Given $x, |x| = n$, at real time, identify the set 
$F_i \in {\cal F'}=\{F \subseteq {\cal F}\ |\ c(F, n) \le c_0\}$ such that
$\forall_{F \in {\cal F'}} \alpha\times a(F_i) \ge a(F)$ and return ${\cal M}(F_i)(x)$.

%$F_i \subseteq {\cal F}$, such that $c(F_i, n_0) \leq c_0$, 
%and from among all $F_j$ such that $c(F_j, n_0) \leq c_0$,
%$\alpha a(F_i) \geq a(F_j)$, and return ${\cal M} (F_i) (x)$.
\end{compactitem}
\end{problem}

%We denote the modified versions of these problems {\sc $\alpha$-Approx-Prob-General}, 
%{\sc $\alpha$-Approx-Prob-$n$-Fixed}, {\sc $\alpha$-Approx-Prob-$c_0$-Fixed}, respectively.
% For the rest of the text, we will focus on the $\alpha$-relaxed versions;
% however, our discussion will apply equally well to the $\alpha = 1$ case (
% which are precisely the three problems described above.)

\stitle{Reusing Skyline Computation is Incorrect:} 
We now argue that it is not sufficient to simply compute
the skyline of classification models for a fixed item size and 
employ that skyline for all $n, c$.
For the following discussion, we focus on $\alpha = 1$; the general $\alpha$ case
is similar.
Given a fixed $n=n_0$, we define the {\em $n_0-$skyline} as the 
set of all feature sets that are {\em undominated} in
terms of cost and accuracy (or equivalently, error, which is $1 -$ accuracy).
A feature set $F_i$ is undominated
if there is no feature set $F_j$,
where
$c(F_j, n_0) < c(F_i, n_0)$ and $a(F_i) \leq a(F_j)$,
and no feature set $F_j$
where 
$c(F_j, n_0) \leq c(F_i, n_0)$ and $a(F_i) < a(F_j)$\footnote{Notice that the $\leq$ operator is placed in different clauses in the two statements.}.
A naive strategy is to enumerate each feature set $F \subseteq {\cal F}$, and
characterize each $F$ by its cross-validation accuracy and average extraction cost over the training dataset.
Once the feature sets are charcterized, iterate through them by increasing cost, and keep
the feature sets whose accuracy is greater than any of the feature sets preceeding it.
The resulting set is the $n_0-$skyline.
However, it is every expensive to enumerate and characterize all feature sets
(especially when the number of features is large), and one of our key contributions
will be to avoid this exhaustive enumeration.
But for the purposes of discussion, let us assume that 
we have the $n_0-$skyline computed.
Note that the skyline feature sets are precisely the ones
we need to consider as possible solutions during real-time
classification for $n = n_0$ for various
values of $c_0$. 

\begin{figure}[h]
\vspace{-10pt}
\centering
\subfigure{\label{fig:tradeoff} \includegraphics[width = 1.3in]{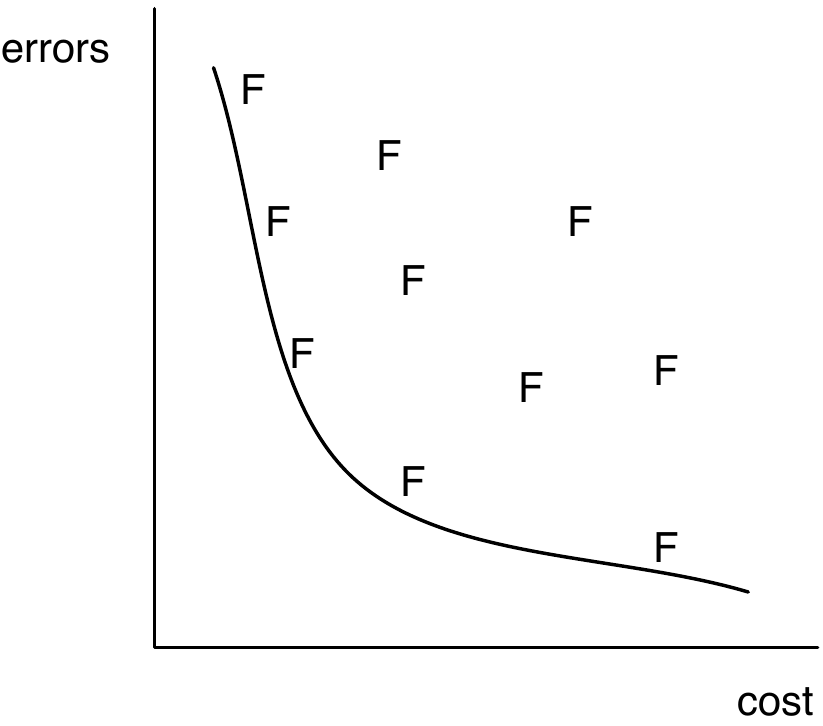}}
\subfigure{\label{fig:scale} \includegraphics[width=1.3in]{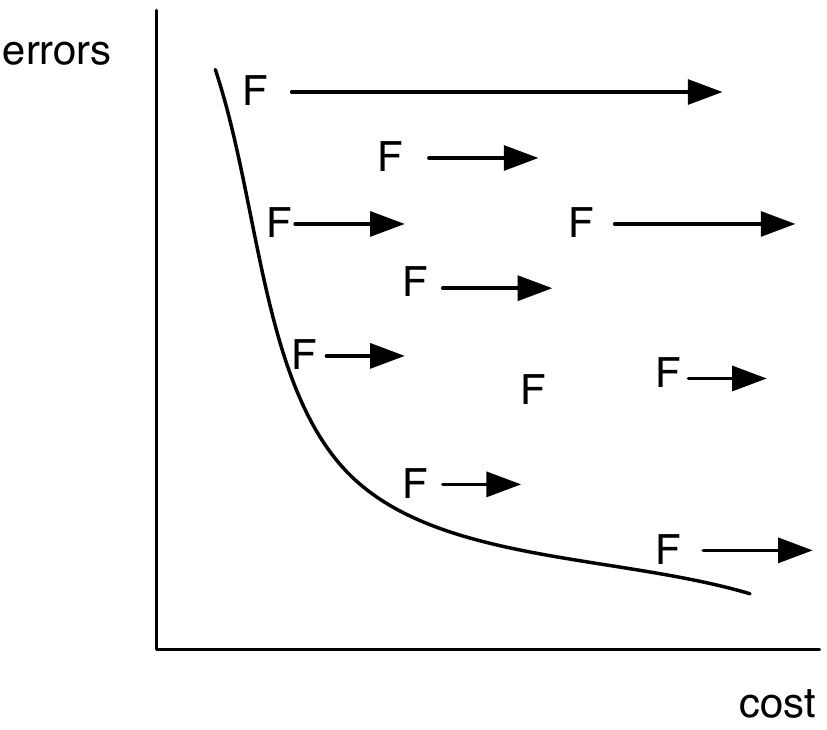}}
\vspace{-5pt}
\caption{(a) Offline computation of all $F$. (b) Points skew across the cost axis as $n$ changes.\label{fig:tradeoffandscale}
}
\vspace{-10pt}
\end{figure}

Then, one approach to solving Problem~\ref{prob} 
could be to simply reuse the $n_0$-skyline for other $n$. 
However, this approach is incorrect, as depicted by
Figure~\ref{fig:tradeoffandscale}.
Figure~\ref{fig:tradeoff} depicts the cost and error of each feature set and the error vs. cost skyline
curve for $n = n_0$,
while Figure~\ref{fig:scale} depicts what happens when 
$n$ changes from $n_0$ to a larger value $n_1$ 
(the same holds when the value changed to a smaller value):
as can be seen in the figure, 
different feature sets move by different amounts,
based on the cost function $c(F_i, n)$.
This is because different polynomials behave differently
as $n$ is varied.
For instance, $2n^2$ is less than $3n + 32$ when $n$ is small, however 
it is significantly larger for big values of $n$.
As a result, a feature set which was on the skyline may now
no longer be on the skyline, and another
one that was dominated could suddenly become part of the
skyline. 

\stitle{Learning Algorithm Properties:} 
We now describe a property about machine learning algorithms
that we will leverage in subsequent sections. 
This property holds because even in the worst case, adding
additional features simply gives us no new useful information that can help us in
classification.
\begin{axiom}[Information-Never-Hurts]\label{ax:gain}
If $F_i \subset F_j$, then $a(F_i) \leq a(F_j)$
\end{axiom}
While this property is known by the machine learning community
to be anecdotally true~\cite{domingos2012few},
we experimentally validate this in our experiments.
In fact, even if this property is violated
in a few cases, our \pdom algorithm can be made more robust by 
taking that into account, as we will see in Section~\ref{sec:cand-set}

%!TEX root=../main.tex

\section{{\bf {\Large \pdom}} Solution}\label{sec:sol}
Our solution follows three steps: 
\begin{compactitem}
\item {\em Feature Set Pruning:}
First, we will start by constructing what we call
as a {\em candidate set}, that is, the set of all feature sets
(and corresponding machine learning models) 
that will be solutions to Problem~\ref{prob}. 
As a side effect, we will find a solution to Problem~\ref{prob-n}.
The candidate set will be a carefully constructed
superset of the skyline feature sets, so that we
do not discard any feature sets that could be useful for any $n$.
We will describe this in Section~\ref{sec:cand-set}.
\item {\em Polydom Index Construction:}
In Section~\ref{sec:poly-dom}, we describe a new data-structure for Problem~\ref{prob}, called the
{\em poly-dominance index}, which compactly represents the candidate
set and allows it to be indexed into given a specific item size $n$ and budget $c$ 
during query time.
In particular, we would like to organize the candidate set so that
it can be efficiently probed even for large candidate sets size.

%even though the candidate set may be large, 
%and the value $n$ is provided during query time, 
%we would like to organize and maintain the candidate
%sets such that it can be efficiently retrieved given $n_0, c_0$.
\item {\em Online Retrieval:}
Lastly, we describe how the {\em poly-dominance index} 
is accessed during query time in Section~\ref{sec:online}.
\end{compactitem}

\subsection{Offline: Constructing the Candidate Set}\label{sec:cand-set}

\begin{figure}

\centering
\begin{tikzpicture}[scale=0.6]
  \node (max) at (0,4) {$\underbrace{{\cal F}^{\star\dagger}}_{0.96}$};
  \node (a1) at (-3,2) {$\underbrace{f_1 f_2 f_3^{\star\dagger}}_{0.86}$};
  \node (a2) at (-1,2) {$\underbrace{f_1 f_2 f_4^{\star\dagger}}_{0.90}$};
  \node (a3) at (1,2) {$\underbrace{f_1 f_3 f_4^{\star\dagger}}_{0.83}$};
  \node (a4) at (3,2) {$\underbrace{f_2 f_3 f_4^{\star\dagger}}_{0.75}$};
  \node (b1) at (-5,0) {$\underbrace{f_1 f_2^{\star\dagger}}_{0.86}$};
  \node (b2) at (-3,0) {$\underbrace{f_1 f_3^\star}_{0.80}$};
  \node (b3) at (-1,0) {$\underbrace{f_1 f_4^\star}_{0.78}$};
  \node (b4) at (1,0) {$\underbrace{f_2 f_3}_{0.75}$};
  \node (b5) at (3,0) {$\underbrace{f_2 f_4^\star}_{0.75}$};
  \node (b6) at (5,0) {$\underbrace{f_3 f_4}_{0.75}$};
  \node (c1) at (-3,-2) {$\underbrace{f_1^{\star\dagger}}_{0.78}$};
  \node (c2) at (-1,-2) {$\underbrace{f_2^{\star\dagger}}_{0.7}$};
  \node (c3) at (1,-2) {$\underbrace{f_3^{\star\dagger}}_{0.75}$};
  \node (c4) at (3,-2) {$\underbrace{f_4^{\star\dagger}}_{0.6}$}; 
  \node (min) at (0,-4) {$\underbrace{\{\}^{\star\dagger}}_{0.5}$};
  
  % \node (b) at (0,2) {$(1,0,1)$};
  % \node (c) at (2,2) {$(1,1,0)$};
  % \node (d) at (-2,0) {$(0,0,1)$};
  % \node (e) at (0,0) {$(0,1,0)$};
  % \node (f) at (2,0) {$(1,0,0)$};
  % \node (min) at (0,-2) {$(0,0,0)$};
  % % \draw (min) -- (d) -- (a) -- (max) -- (b) -- (f)
  % (e) -- (min) -- (f) -- (c) -- (max)
  % (d) -- (b);
   \draw  (max) -- (a1) -- (b1) -- (c1) -- (min);
   \draw  (min) -- (c2) -- (b1) -- (a2) -- (max);
   \draw (min) -- (c3) -- (b2) -- (a3) -- (max);
   \draw (min) -- (c4) -- (b6) -- (a4) -- (max);
   \draw  (b3) -- (c1) -- (b2);
   \draw (b4) -- (c2) -- (b5);
   \draw (b4) -- (c3) -- (b6);
   \draw (b3) -- (c4) -- (b5);
   \draw (a1) -- (b2);
   \draw (a2) -- (b3) -- (a3);
   \draw (a1) -- (b4) -- (a4);
   \draw (a2) -- (b5) -- (a4);
   \draw (a3) -- (b6);

\end{tikzpicture}
\caption{Lattice: $\star$ indicates the nodes that will be expanded for $\alpha = 1$; 
$\dagger$ indicates the nodes that will be expanded for $\alpha = 1.1$}\label{fig:lattice}
\vspace{-10pt}
\end{figure}
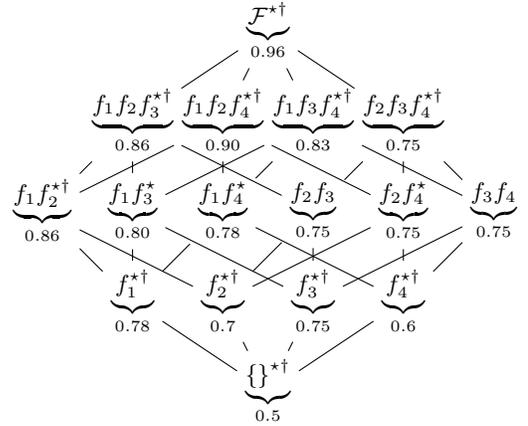

We will construct the candidate set using a bi-directional
search on the lattice of all subsets of features, depicted in Figure~\ref{fig:lattice}.

When a sequence of features is listed, this sequence
corresponds to the feature set containing those features.
In the figure, feature sets are listed along with their accuracies 
(listed below the feature set)\footnote{We have chosen
accuracy values that satisfy Axiom~\ref{ax:gain} in the previous section.}.
An edge connect two features sets that differ in one feature.
For now, ignore the symbols $\star, \dagger$, we will
describe their meaning subsequently. 
The feature set corresponding to $\cal F$ is 
depicted at the top of the lattice,
while the feature set corresponding to the 
empty set is depicted at the bottom.
The feature sets in between have $0$ to $|\cal F|$
features.
In the following we use feature sets and nodes in the
lattice interchangably.

\stitle{Bidirectional Search:}
At one extreme, we have the empty set $\{\}$, and at the other
extreme, we have ${\cal F}$. 
We begin by learning and characterizing 
the best machine learning model for $F = \{\}$,
and for $F = {\cal F}$: i.e., we learn the best machine learning model, 
represented as ${\cal M}(F)$, and learn the 
accuracy $a(F)$ (listed below the node)
and $c(F, n)$ for the model\footnote{Note that the cost of a model (i.e. featureset)
is simply the sum of the individual features and can re-use previously computed costs.}
We call this step {\em expanding} a feature set,
and a feature set thus operated on is called an {\em expanded}\
feature set.

At each round, we expand the feature sets in the next layer, in both
directions. 
We stop once we have expanded all nodes.
In our lattice in Figure~\ref{fig:lattice}, 
we expand the bottom and top layer each consisting of 1 node,
following which, we expand the second-to-bottom layer consisting of 4 nodes,
and the second-to-top layer again consisting of 4 nodes, 
and then we finally expand the middle layer consisting of 6 nodes.

However, notice that the total number of nodes in the lattice
is $2^{|\cal F|}$, and even for relatively small ${\cal F}$,
we simply cannot afford to expand all the nodes in the lattice. 
Therefore, we develop pruning conditions to avoid
expanding all the nodes in the lattice. 
Note that all the pruning conditions we develop are guaranteed
to return an accurate solution.
That is, we do not make approximations at any point that
take away the optimality of the solution.

\stitle{Dominated Feature Sets:} We now define what we mean
for a feature set to dominate another.
\begin{definition}[dominance]
A feature set $F_i$ dominates a feature set $F_j$ 
if $\forall n,$ $c(F_i, n)$ $\leq c(F_j, n)$ and
$\alpha \times a(F_i) \geq a(F_j)$
\end{definition}
As an example from Figure~\ref{fig:lattice},
consider node $f_3 f_4$ and $f_2 f_3 f_4$ on the right-hand
extreme of the lattice: the accuracies of both these
feature sets is the same, while the
cost of $f_2 f_3 f_4$ is definitely higher
(since an additional feature $f_2$ is evaluated).
Here, we will always prefer to use $f_3 f_4$ over
$f_2 f_3 f_4$, and as a result,
$f_2 f_3 f_4$ is dominated by $f_2 f_3$.

Overall, a feature set $F_j$ that is dominated is simply not under consideration 
for any $n$, because it is not going to be the solution to
Problems~\ref{prob}, \ref{prob-n}, or \ref{prob-c0}, given
that $F_i$ is a better solution.
We formalize this as a theorem:
\begin{theorem}\label{thm:dominated-solution}
A feature set that is dominated cannot be the solution to
Problems~\ref{prob}, \ref{prob-n}, or \ref{prob-c0} for
any $n$.
\end{theorem}
Given the property above, we need to find domination rules
that allow us to identify and discard dominated feature sets.
In particular, in our lattice, this corresponds to not
expanding feature sets.

\stitle{Pruning Properties:}
Our first property dictates that we should not expand
a feature set that is strictly ``sandwiched between'' two
other feature sets. 
It can be shown that any such feature set is
dominated, and therefore, using Theorem~\ref{thm:dominated-solution},
can never be a solution to any of the problems listed
in the previous section.  
\begin{property}[Sandwich-Property]\label{prop:sandwich}
If $F_i  \subset F_k$, and $\alpha \times a(F_i) \geq a(F_k)$,
then no $F_j$ such that $F_i \subset F_j \subset F_k$, needs to be
expanded.
\end{property}
Intuitively, if there is a feature set $F_i$ that dominates an $F_j$,
while $F_i \subset F_j$, then all other feature sets betweeen $F_i$
and $F_j$ are also dominated.
Consider Figure~\ref{fig:lattice}, with $\alpha = 1$,
let $F_i = f_3$, and $F_k = f_2 f_3 f_4$,
since $a (F_i) = a (F_k)$, feature sets
$F_j$ corresponding to $f_2 f_3$ and $f_3 f_4$ 
need not be expanded: in the figure, 
both these feature sets have precisely the same
accuracy as $f_3$, but
have a higher cost.

In Figure~\ref{fig:lattice} once again for $\alpha = 1$,
let us consider how many expansions the previous property saves us
while doing bidirectional search:
We first expand ${\cal F}$ and $\{\}$,
and then we expand all nodes in the second-to-top layer
and the second-to-bottom layer.
Then, from the next layer, $f_2 f_3$ and $f_3 f_4$ will not be
expanded 
(using the argument from the previous
paragraph), while the rest are expanded.
Thus, we save two expansions. 
The expanded nodes in the lattice are denoted using $\star$s.

Now, on changing $\alpha$ slightly, the number of evaluations goes
down rapidly. 
The nodes expanded in this case are denoted using a $\dagger$.
Let us consider $\alpha = 1.1$.
Once again, nodes in the top two and bottom two layers are expanded.
However, only $f_1 f_2$ in in the middle layer needs to be expanded.
This is because:
\begin{compactitem}
\item $f_1 f_3$ and $f_1 f_4$ are sandwiched between $f_1$ and $f_1 f_3 f_4$
\item $f_2 f_3$ and $f_2 f_4$ are sandwiched between $f_2$ and $f_2 f_3 f_4$
\item $f_2 f_3$ and $f_3 f_4$ are sandwiched between $f_3$ and $f_2 f_3 f_4$
\end{compactitem}
The previous property is hard apply directly (e.g., before expanding
every feature set we need to verify if there exists a pair of feature
sets that sandwich it). Next, we describe a property that specifies
when it is safe to stop expanding all non-expanded ancestors of a
specific node.

\begin{property}[Covering-Property]
If $F_i  \subset F_{k_i}, \forall i \in 1 \ldots r$
such that $\cup_{i\in 1\ldots r} F_{k_i} = {\cal F}$, and
$\alpha \times a(F_i) \geq a(F_{k_i}),  \forall i \in 1 \ldots r$, then no feature set sandwiched
between $F_i$ and $F_{k_i}$ needs to be expanded.
\end{property}

This property states if any set of feature sets $F_{k_i}$ 1) contain $F_i$, 2) in
aggregate covers all the features in ${\cal F}$ and 3) are dominated
by $F_i$, then all feature sets between $F_i$ and $F_{k_i}$ do not need to be expanded.  
The inverse property for pruning descendents also holds.

We use this property to extend the bidirectional search with an
additional pruning step.  Let the top frontier ${\cal F}_{top}$ be
the set of feature sets expanded from the top for which no child
feature set has been expanded, and let ${\cal F}_{bot}$ be similarly
defined from the bottom.  By directly applying the {\it Covering-Property},
we can prune the parents of $F \in {\cal F}_{bot}$ if $\forall F'
\in {\cal F}_{top}\ F \subseteq F'$, $F$ dominates $F'$.  We can
similarly use the inverse of the property to prune feature sets in
the top frontier.

% if there exists a set of $F_j$s that each contain $F_i$
%and in aggregate cover all the features in ${\cal F}$, while being dominated by $F_i$,
%then you can stop expanding descendants of $F_i$, since they will
%be dominated by $F_i$ as well.
%The basic intuition is that that no matter which feature you add to $F_i$, you 
%end up in a feature set dominated by $F_i$.
%The second property is similar, but in the inverse direction.

%develop two specializations of the sandwich property
%that are actually easier to check.
%In fact, these two properties allow us to stop expanding any node that is an
%ancestor or a descendant of a specific node.
%
%\begin{property}[Reverse-Covering-Property]
%If $F_i  \supset F_{k_i}, i \in 1 \ldots r$, such that $\cup_i F_{k_i} = {\cal F}$,
%and $\alpha \times a(F_i) \leq a(F_{k_i}), \forall i \in 1 \ldots r$, then
%no $F_j$ such that $F_i \subset F_j$, needs to be expanded.
%\end{property}
%\agp{Candidate for deletion ends}

\stitle{Properties of Expanded Nodes:}
We have the following theorem, that is a
straightforward consequence of Property~\ref{prop:sandwich}:
\begin{theorem}
The set of expanded nodes form a superset of the 
skyline nodes for any $n$.
\end{theorem}
In figure~\ref{fig:lattice}, the set of expanded nodes
(denoted by $\star$ for $\alpha = 1$ and $\dagger$ for $\alpha = 1.1$)
are the ones relevant for any $n$.

\stitle{Candidate Nodes:}
Given the expanded set of nodes, 
two properties to allow us to prune away some of the expanded
but dominated nodes to
give the {\em candidate} nodes. 
Both these properties are straightforward consequences of
the definition of dominance.
\begin{property}[Subset Pruning-Property]
If $F_i  \subset F_k$, and $\alpha \times a(F_i) \leq a(F_k)$,
then $F_k$ does not need to be retained as a {\em candidate}
\end{property}
For instance, even though $f_2 f_3 f_4$ and $f_3$
are both expanded, $f_2 f_3 f_4$ does not need to be
retained as a candidate node when $f_3$ is present
(for any $\alpha$);
also, $f_1 f_3 f_4$ does not need to be retained
as a candidate node when $f_1$ is present for $\alpha = 1.1$.

The next property is a generalization of the previous,
when we have a way of evaluating polynomial dominance.

\begin{property}[Poly-Dom Pruning-Property]
If $c(F_i, x)$ $ < c(F_k,x), \forall x \geq 0$, 
and $\alpha \times a(F_i) \leq a(F_k)$,
then $F_k$ does not need to be retained as a {\em candidate}
\end{property}
The next theorem states that we have not made
any incorrect decisions until this point, i.e.,
the set of candidate nodes includes all the nodes
that are solutions to Problems~\ref{prob}, \ref{prob-n}, \ref{prob-c0} for
all $n$.
\begin{theorem}
The set of candidate nodes form a superset of the 
skyline nodes for any $n$.
\end{theorem}
% \agp{Can we prove that every node in the candidate
% set will be the solution to some $n, c_0$?}

% \agp{Can we prove that under certain assumptions about
% $a$, $c$, the number of nodes expanded is small,
% and the number of candidates nodes is small as well?}

\stitle{Algorithm:} The pseudocode for
the algorithm can be found in the appendix split into:
Algorithm~\ref{alg:expand-enumerate} 
(wherein the lattice is traversed and the nodes are expanded)
and Algorithm~\ref{alg:candidate} 
(wherein the dominated expanded nodes are removed to give
the candidate nodes).

In brief, Algorithm~\ref{alg:expand-enumerate} maintains
two collections: $\sf{frontierTop}$ and $\sf{frontierBottom}$,
which is the frontier (i.e., the boundary) of already expanded nodes
from the top and bottom of the lattice respectively.
The two collections $\sf{activeTop}$ and $\sf{activeBottom}$
contain the next set of nodes to be expanded.
When a node is expanded, its children in the lattice are added to 
$\sf{activeTop}$ if the node is expanded ``from the top'',
while its parents in the lattice are added to $\sf{activeBottom}$
if the node is expanded ``from the bottom''.

Note that there may be smart data structures we could use
to check if a node is sandwiched or not, or
when enumerating the candidate set.
Unfortunately, the main cost is dominated by the
cost for expanding a node 
(which involves training a machine learning
model given a set of features and estimating its
accuracy),
thus these minor improvements do not
improve the complexity much.

%\begin{quote}
%{\sc Expand-Enumerate}(${\cal F}, n_0, \alpha$)

%\end{quote}

% \stitle{Special Case for $n = n_0$:} 
% Need to construct skyline for $n = n_0$
% \begin{theorem} The following relationship holds:
% The skyline for $\forall n_0, n = n_0 \subseteq$ candidate nodes $\subseteq $ expanded nodes.
% \end{theorem}

% \begin{theorem} Setting candidate nodes to be
% equal to the skyline nodes is incorrect.
% \end{theorem}
% \begin{proof}
% counterexample...
% \end{proof}

\begin{figure}[h!]
\vspace{-10pt}
\centering
\includegraphics[width = 3in]{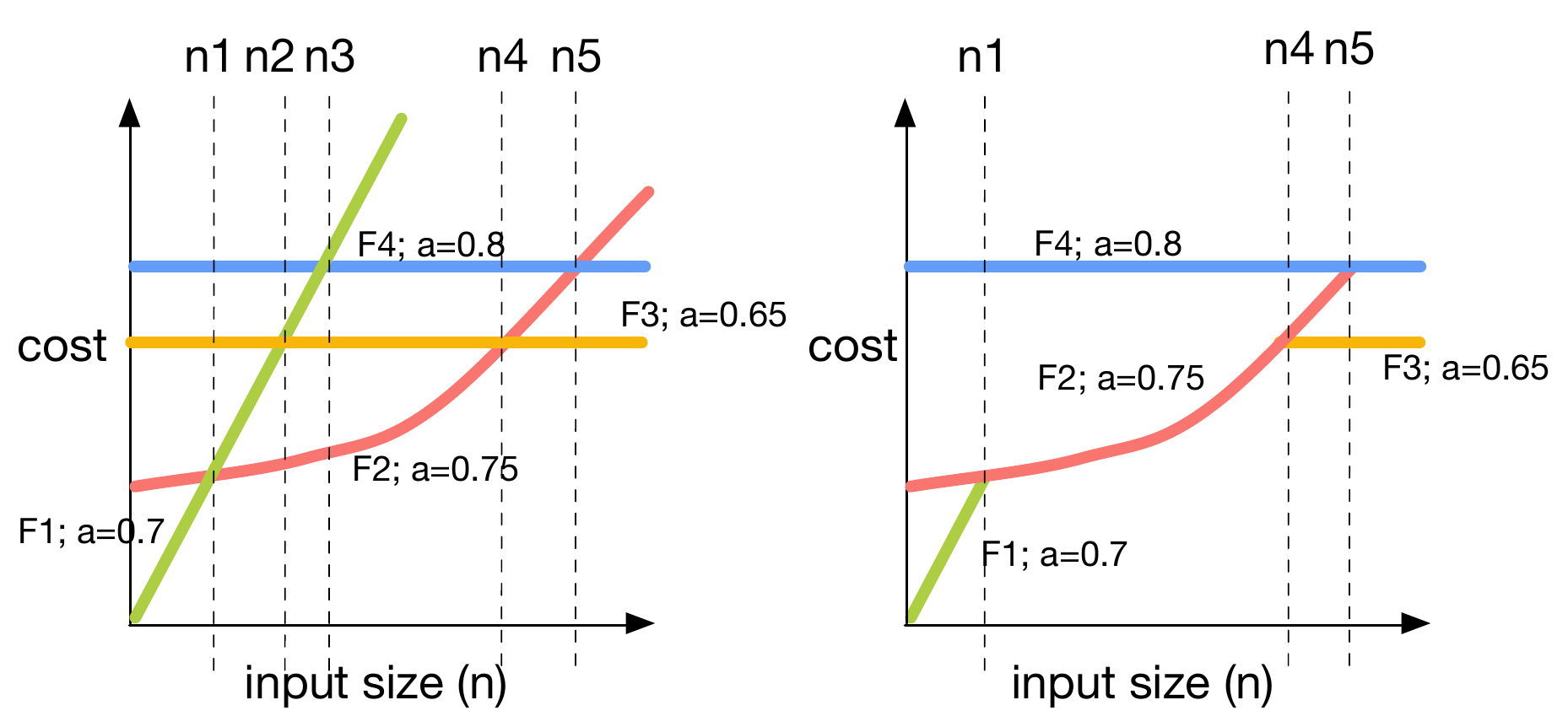}
\caption{Incremental Poly-Dom Creation}\label{fig:poly-dom}
\vspace{-10pt}
\end{figure}

\stitle{Discussion:}
When the number of features in \calF\ 
are in the thousands, the lattice would be massive.
In such cases, even the number of expanded nodes can be in the 
millions. 
Expanding each of these nodes can take a significant time, since
we would need to run our machine learning algorithm on each 
node (i.e., feature set).
In such a scenario, we have two alternatives:
(a) we apply a feature selection algorithm~\cite{saeys2007} that
allows us to bring the number of features under consideration
to a smaller, more manageable number, or;
(b) we apply our pruning algorithm in a progressive modality.
In this modality, a user provides
a precomputation pruning cost budget, and the pruning algorithm 
picks the ``best $\alpha$'' to meet the precomputation
cost budget
(i.e., the smallest possible $\alpha$ for which
we can apply the lattice pruning algorithm within 
the precomputation cost budget.) 
The approach is the following: we start with a large $\alpha$ 
(say 2),
and run the lattice pruning algorithm. 
Once complete, we can reduce $\alpha$ by a small amount,
and rerun the lattice pruning algorithm, and so on,
until we run out of the precomputation cost budget.
We can make use of the following property:
\begin{property}
The nodes expanded for $\alpha_1, 1 \le \alpha_1 < \alpha_2 \le 2$ is
a superset of the nodes expanded for $\alpha_2$.
\end{property}
Thus, no work that we do for larger $\alpha$s are wasted
for smaller $\alpha$s: as a result,
directly using the $\alpha$ that is the best for
the precomputation cost budget would be equivalent to
the above procedure, since the above procedure
expands no more nodes than necessary.

\stitle{Anti-Monotonicity:}
Note that there may be practical scenarios 
where the assumption of monotonicity,
i.e., Axiom~\ref{ax:gain}, does not hold,
but instead, a relaxed version of monotonicity holds,
that is, 
\begin{axiom}[Information-Never-Hurts-Relaxed]\label{ax:gain-relaxed}
If $F_i \subset F_j$, then $a(F_i) \leq a(F_j) + e$
\end{axiom}
Here, if $F_i$ is a subset of $F_j$, then $a(F_i)$ cannot be
larger than $a(F_j) + e$.
Intuitively, the violations of monotonicity, if any are 
small---smaller than $e$ (we call this the {\em error} 
in the monotonicity.)
Note that when $e = 0$, we have Axiom~\ref{ax:gain} once again.

In such a scenario, only the lattice construction
procedure is modified by ensuring that we do not prematurely
prune away nodes (say, using the sandwich property) that can still be 
optimal.
We use the following modified sandwich property:
\begin{property}[Sandwich-Property-Relaxed]\label{prop:sandwich-relaxed}
If $F_i  \subset F_k$, and $\alpha \times (a(F_i) - e) \geq a(F_k)$,
then no $F_j$ such that $F_i \subset F_j \subset F_k$, needs to be
expanded.
\end{property}
With the above property, we have a more stringent condition,
i.e., that $\alpha \times (a(F_i) - \delta)$ and not simply 
$\alpha \times a(F_i)$ has to be greater than $a(F_k)$.
As a result, 
fewer pairs of nodes $i, k$ qualify,
and as a result, fewer nodes $j: F_i \subset F_j \subset F_k$
are pruned without expansion.

Subsequently, when deciding whether to remove
some of the expanded nodes to give candidate nodes, we
have the true accuracies of the expanded nodes, we no longer
need to worry about the violations of monotonicity.

\subsection{Offline: Constructing the Index}\label{sec:poly-dom}

We begin by collecting the set of candidate nodes from the previous step.
We denote the set of candidate nodes as ${\cal C}, |{\cal C}| = k$.
We now describe how to construct the {\em poly-dom} index.

\stitle{Alternate Visualization:} 
Consider an alternate way of visualizing the set of candidate nodes,
depicted in Figure~\ref{fig:poly-dom}(left).
Here, we depict the cost $c(F_i,n)$ for each of the candidate nodes,
as a function of $n$. 
Also labeled with each cost curve is the accuracy. 
Recall that unlike cost, the accuracy 
stays constant independent of the input size $n$.
We call each of the curves corresponding 
to the candidate nodes as {\em candidate curves}.
We depict in our figure four candidate curves,
corresponding to feature sets $F_1, F_2, F_3, F_4$.
In the figure, we depict five `intersection points',
where these candidate curves cross each other.
We denote, in ascending order, the intersection points,
as $n_1 < \ldots < n_r$. 
In Figure~\ref{fig:poly-dom}(left), $r = 5$.
It is easy to see that the following holds:
\begin{lemma}
For all intersection points $n_1 < \ldots < n_r$ 
between candidate curves,
storing the skyline for the following ranges 
$(-\infty, n_1), $ $[n_1, n_2), $ $ \ldots, $ $[n_{r-1}, n_r),$ $[n_r, \infty) $
is sufficient for Problems~\ref{prob},\ref{prob-n}, \ref{prob-c0}, since
for any such range, the skyline is fixed.
\end{lemma}
For $(-\infty, n_1)$, $F_4$ has accuracy 0.8, $F_3$ has 
accuracy 0.65, $F_2$ has accuracy 0.76, and $F_1$ has accuracy 0.7.
The skyline of these four candidate sets for $n < n_1$ is
$F_4, F_2, F_1$; $F_3$ is dominated by $F_2$ and $F_1$ both of which have
lower cost and higher accuracy.

The lemma above describes the 
obvious fact that the relationships between
candidate curves (and therefore nodes) do not change between the
intersection points, and therefore, we only need to record what changes
at each intersection point.
Unfortunately, with $r$ candidate curves, there can be as many as $r^2$ intersection points.

Thus, we have a naive approach to compute the index that allows us to 
retrieve the optimal candidate curve for each value of $n_0, c_0$:
\begin{compactitem}
\item for each range, we compute the skyline of candidate nodes, 
and maintain it ordered on cost
\item when $n_0, c_0$ values are provided
at query time, we perform a binary search
to identify the appropriate range for $n_0$,
and do a binary search to identify the candidate node that
that respects the condition on cost. 
\end{compactitem}
Our goal, next, is to identify ways to prune the number of 
intersection points so that we do not need to index and maintain
the skyline of candidate nodes for many intersection points.

\stitle{Traversing Intersection Points:}
Our approach is the following:
We start with $n = 0$, and order the curves at that point in terms of cost.
We maintain the set of curves in an ordered fashion throughout.
Note that the first intersection point after the origin
between these  curves  (i.e., the one that has smallest $n, n>0$)
has to be an intersection of two curves that 
are next to each other in the ordering at $n = 0$.
(To see this, if two other curves intersected that were not adjacent,
then at least one of them would have had to intersect
with a curve that is adjacent.)
So, we compute the intersection points for all pairs of adjacent curves
and maintain them in a priority queue (there are at most $k$ intersection points).

We pop out the smallest intersection point from this priority queue.
If the intersection point satisfies certain conditions (described below),
then we store the skyline for that intersection point.
We call such an intersection point an {\em interesting intersection point}.
If the point is not interesting, we do not need to store the skyline for
that point.
Either way, when we have finished processing this intersection point, we do the following:
we first remove the intersection point from the priority queue.
We then add two more intersection points to the priority queue, 
corresponding to the intersection points with the new
neighbors of the two curves that intersected with each other.
Subsequently, we may exploit the property that 
the next intersection point 
has to be one from the priority queue of intersection points
of adjacent curves. 
We once again pop the next intersection point
from the priority queue and the process continues.

The pseudocode for our procedure is listed in Algorithm~\ref{alg:poly-dom-int} in the appendix.
The array ${\sf sortedCurves}$ records the candidate curves sorted on cost,
while the priority queue ${\sf intPoints}$ contains the 
intersection points of all currently adjacent curves.
As long as ${\sf intPoints}$ is not empty, we keep 
popping intersection points from it, update ${\sf sortedCurves}$ to 
ensure that the ordering is updated, and 
add the point to the list of skyline recomputation points if the
point is an interesting intersection point.
Lastly, we add the two new intersections points of the 
curves that intersected at the current point.

\begin{figure}
\vspace{-10pt}
\centering
\hspace{.4in}\includegraphics[width=2in]{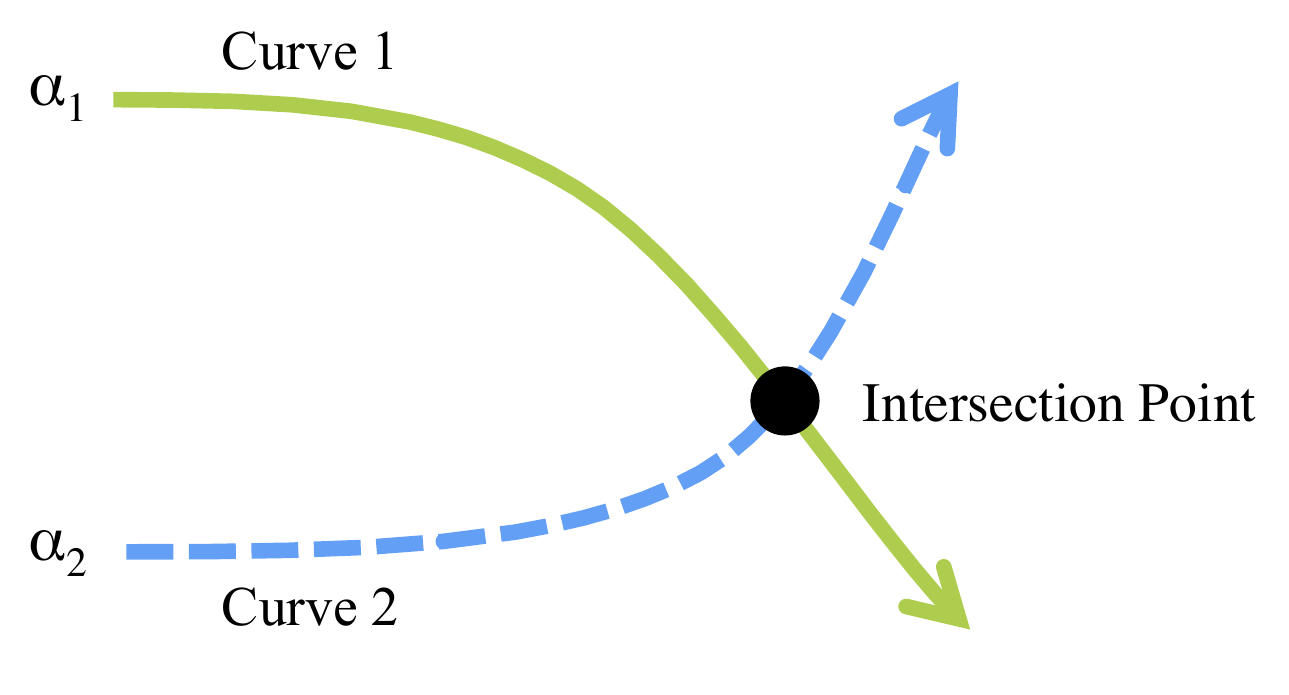}
%\begin{tikzpicture}[node distance=1cm, auto,]
% %nodes
% \node[auto, align=left] (market) {curve 1\\
%accuracy a};
% \node[auto, inner sep=5pt,below=0.5cm of market, align=left] (formidler) {curve 2 \\ accuracy b};
% % We make a dummy figure to make everything look nice.
% \node[below=0.25cm of market] (dummy) {};
% \node[right=2cm of dummy] (t) {Intersection Point}
%   edge[pil,<-,bend right=45] (market.east) 
%   edge[pil, <-, bend left=45] (formidler.east); 
%% \node[left=of dummy] (g) {Ultimate lender}
% %  edge[pil, bend right=45] (market.west)
%  % edge[pil, bend right=45] (formidler.west)
% %  edge[pil,<->, bend left=45] node[auto] {Direct (a)} (t);
%\end{tikzpicture}
\caption{Reasoning about Intersection Points}\label{fig:intersection}
\vspace{-10pt}
\end{figure}

\stitle{Pruning Intersection Points:}
Given a candidate intersection point, we
need to determine if we need to store the skyline 
for that intersection point.
We now describe two mechanisms we use to prune away
``uninteresting'' intersection points.
First, we have the following theorem, which uses
Figure~\ref{fig:intersection}:
\begin{theorem}\label{thm:intersection}
We assume that for no two candidate nodes, the accuracy is same.
The only intersection points (depicted in Figure~\ref{fig:intersection},
where we need to recompute the skyline are
the following:
\begin{compactitem}
\item Scenario 1: Curve 1 and 2 are both on the skyline, and $\alpha_1 > \alpha_2$.
In this case, the skyline definitely changes, and therefore
the point is an interesting intersection point.
\item Scenario 2: Curve 1 is not on the skyline while Curve 2 is.
Here, we have two cases:
if $\alpha_1 > \alpha_2$ then the
skyline definitely changes, and
if $\alpha_1 < \alpha_2$, then the skyline changes 
iff there is no curve below Curve 2, whose accuracy is greater than $\alpha_2$.
\end{compactitem}
\end{theorem}

As an example of how we can use the above theorem, consider
Figure~\ref{fig:poly-dom},
specifically, intersection point $n_1$ and $n_2$.
Before $n_1$, the skyline was $[F_1, F_2, F_4]$, with 
$F_3$ being dominated by $F_1, F_2$.
At $n_1$, $F_1$ and $F_2$ intersect. 
Now, based on Theorem~\ref{thm:intersection}, since 
the lower curve (based on cost) before the intersection has lower accuracy,
the curve corresponding to $F_2$ now starts
to dominate the curve corresponding to $F_1$,
and as a result, the skyline changes. Thus, this intersection
point is indeed interesting.

Between $n_1$ and $n_2$, the skyline was $[F_2, F_4]$, since $F_3$ and $F_1$ are both
dominated by $F_2$ (lower cost and higher accuracy).
Now, at intersection point $n_2$, curves $F_3$ and $F_1$ intersect.
Note that neither of these curves are on the skyline.
Then, based on Theorem~\ref{thm:intersection}, 
we do not need to recompute the skyline for $n_2$.

Recall that we didn't use $\alpha$ approximation at all.
Since we already used $\alpha$ to prune candidate
nodes in the first step, we do not use it again to prune
potential curves or intersection points, since that may
lead to incorrect results.
In our experience, the lattice pruning step
is more time-consuming (since we need to train
a machine learning model for each expanded node),
so it is more beneficial to use $\alpha$ in
that step.
We leave determining how to best integrate $\alpha$ into the learning algorithms
as future work.  Finally, the user can easily integrate domain knowledge, such as the distribution of
item sizes, into the algorithm to further avoid
computing and indexing intemediate intersection
points.

\stitle{Determining the Skyline:}
As we are determining the set of interesting intersection points,
it is also easy to determine and maintain the skyline
for each of these points.
We simply walk up the list of candidate nodes
at that point, sorted by cost, and 
keep all points
that have not been dominated by previous points.
(We keep track of the highest accuracy seen so far.)

% Before $n_1$, we have $[F_1, F_2, F_3, F_4]$ as the sorted order.
% We  maintain a boolean array of the candidate nodes that are part of
% the skyline for $<n_1$. 
% One way to determine the skyline is to sort all the candidate
% nodes for $n_1$ based on cost, and walk up starting from the 
% candidate node with the lowest cost.
% We keep track of the highest accuracy seen so far.
% If, on encountering a candidate node, the accuracy of the 
% candidate node is lower then it is not part of the skyline,
% while otherwise, it is part of the skyline.
% We denote the skyline for $[n_{i-1}, n_i)$ as $sky(n_i)$.
% For the cast of $n_1$, this is $[1, 1, 0, 1]$.

% We also maintain the array of the
% best accuracies before a certain index.
% For before $n_1$, the bestBefore array is $[-, 0.7, 0.75, 0.75]$.

\stitle{Index Construction:}
Thus, our polydom indexing structure is effectively is a two-dimensional
sorted array, where we store 
the skyline for different values of $n$.
We have, first, a sorted array corresponding to the sizes of the
input.
Attached to each of these locations is an array containing the skyline
of candidate curves.

\begin{figure}[h!]
\vspace{-10pt}
\centering
\includegraphics[width = 2in]{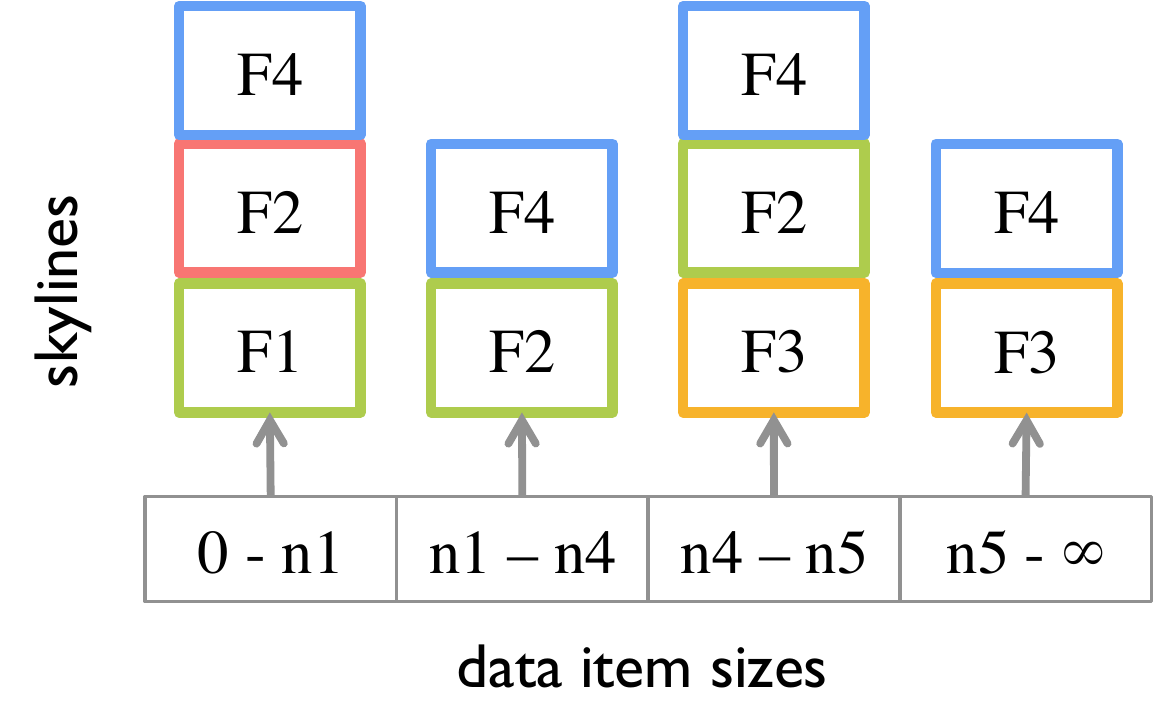}
\caption{Poly-Dom Index for the feature sets in Figure~\ref{fig:poly-dom}}\label{fig:linked-list}
\vspace{-10pt}
\end{figure}

For the intersection curve depicted in Figure~\ref{fig:poly-dom},
the index that we construct is depicted in Figure~\ref{fig:linked-list}.

\if{0}
\agp{I don't know who added this, but I am not sure if we want to do this.}
We should discuss the following indexing strategies and their impact
on quality/performance:

\begin{compactitem}
\item an index entry at every intersection point
\item deterministically and exponentially spaced out points to index
\item polydom index
\end{compactitem}
\fi

\subsection{Online: Searching the Index} \label{sec:online}
Given an instance of Problem~\ref{prob} at
real-time classification time, 
finding the optimal model is simple,
and involves two steps. We describe these 
steps as it relates to Figure~\ref{fig:linked-list}.

\begin{compactitem}
\item We perform a binary search on the $n_i$ ranges,
i.e., horizontally on the bottom most array, to identify
the range within which the input item size $n_0$ lies.
\item We then perform a binary search on the 
candidate nodes,
i.e., vertically for the node identified in the
previous step,
to find the candidate node for which
the cost is the largest cost is less than the target cost $c$.
Note that we can perform binary search because this
array is sorted in terms of cost.
We then return the model corresponding to
the given candidate node.
\end{compactitem}

Thus, the complexity of searching for
the optimal machine learning model is:
$O(\log t_{int} + \log t_{cand})$:
where $t_{int}$ is the number of interesting intersection points,
while $t_{cand}$ is the number of candidate nodes on the skyline.

\section{{\Large {\bf \gm}} Solution} \label{sec:gm-solution}
The second solution we propose, 
called \gm, 
is a simple adaptation of the technique
from Xu {\em et al.}~\cite{DBLP:conf/icml/XuWC12}.
Note that Xu {\em et al.}'s technique does not apply to
generic machine learning algorithms, and only
works with a specific class of SVMs;
hence we had to adapt it to apply to all machine
learning algorithms as a black box.
Further, this algorithm (depicted in Algorithm~\ref{alg:greedy} in the 
appendix) only works with a single
size; multiple sizes are handed as described subsequently.
For now, we assume that the following procedure
is performed with the median size of items.

\stitle{Expansion:} Offline, the algorithm works as follows:
for a range of values of $\lambda \in {\cal L}$, the
algorithm does the following.
For each $\lambda$,
the algorithm considers adding one feature at a time to the current set of features
that improves the most
the function
\begin{quote}
gain = (increase in accuracy) --- $\lambda \times$ (increase in cost)
\end{quote}
This is done by considering adding one feature at a time,
{\em expanding} the new set of features,
and estimating its accuracy and cost.
(The latter is a number rather than
a polynomial --- recall that
the procedure works on a single size of item.)
Once the best feature is added, the corresponding machine learning model for that set of features
is recorded.
This procedure is repeated until all the features are added,
and then repeated for different $\lambda$s.
The intuition here is that the $\lambda$ dictates
the priority order of addition of features:
a large $\lambda$ means a higher preference for cost,
and a smaller $\lambda$ means a higher preference for accuracy.
Overall, these sequences (one corresponding to every $\lambda \in {\cal L}$)
correspond to a number of {\em depth-first} explorations of the lattice,
as opposed to \pdom, which explored the lattice in a {\em breadth-first} manner.

\stitle{Indexing:} We maintain two indexes,
one which keeps track of the sequence of feature sets expanded for each $\lambda$,
and one which keeps the skyline of accuracy vs. cost for the entire set of feature sets.
The latter is sorted by cost.
Notice that since we focus on a single size, for the latter index, 
we do not need to worry about cost functions, we simply use the cost values
for that size.

\stitle{Retrieval:} Online, when an item is provided, the algorithm performs a binary search
on the skyline, picks the desired feature set
that would fit within the cost budget.
Then, we look up the $\lambda$ corresponding to that model,
and then add features starting from the first feature, computing one feature at a time,
until the cost budget is exhausted for the given item.
Note that we may end up at a point where we have evaluated
more or less features than the feature set we started off with,
because the item need not be of the same size as the item size used
for offline indexing.
Even if the size is different, since we have the entire
sequence of expanded feature sets recorded for each $\lambda$,
we can,
when the size is larger, add a subset of features
and still get to a feature set (and therefore a machine learning model) that is good,
or when the size is smaller, add a superset of features (and therefore a model)
and get to an even better model.

\stitle{Comparison:} This algorithm has
some advantages compared to \pdom:
\begin{compactitem}
\item The number of models expanded is simply $|{\cal L}|\times |{\cal F}|^2$,
unlike \pdom, whose number of expanded models 
could grow exponentially in ${\cal F}$ in the worst case.
\item Given the number of models stored is small (proportional to
$|{\cal F}| \times |{\cal L}|$), the lookup can be 
simple and yet effective.
\item The algorithm is any-time; for whatever reason if a feature evaluation
cost is not as predicted, it can still terminate early with a good model,
or terminate later with an even better model.
\end{compactitem}
\gm also has some disadvantages compared to \pdom:
\begin{compactitem}
\item It does not provide any guarantees of optimality.
\item Often, \gm returns models that are worse than \pdom. 
Thus, in cases where accuracy is crucial, we need to use \pdom.
\item The $\lambda$ values we iterate over, 
i.e., ${\cal L}$, requires hand-tuning,
and may not be easy to set. 
Our results are very sensitive to this cost function.
\item Since \gm uses a fixed size, for items that are of a very different size,
it may not perform so well.
\end{compactitem}
We will study the advantages and disadvantages in our experiments.

%Greedy miser is very sensitive to cost function \agp{this is something eugene said}

\stitle{Special Cases:}  We now describe two special cases of \gm that merit attention:
we will consider these algorithms in our experiments as well.
\begin{compactitem}
\item \gacc: This algorithm is simply \gm where ${\cal L} = \{ -\infty \}$;
that is, this algorithm adds one at a time, the feature with the smallest
cost at the median size.
\item \gcost: This algorithm is simply \gm where ${\cal L} = \{ \infty \}$;
that is, this algorithm adds one at a time, the feature that adds the most accuracy
with no regard to cost.
\end{compactitem}
Note that these two algorithms get rid of one of the disadvantages of \gm,
i.e., specifying a suitable ${\cal L}$.

%!TEX root=../main.tex

\section{Experiments}\label{sec:exp}

Online prediction depends on two separate phases---an offline
phase to precompute machine learning models and data structures,
and an online phase to make the most accurate prediction within a
time budget.  To this end, the goals of our evaluation are three-fold:
First, we study how effectively the \pdom and \gm-based algorithms can
prune the feature set lattice and thus reduce the number of models that need to
be trained and indexed during offline pre-computation.  Second, we study how
these algorithms affect the latency and accuracy of the models that are retrieved online.
Lastly, we verify the extent to which our anti-monotonicity assumption holds in real-world datasets.

To this end, we first run extensive simulations to understand the
regimes when each algorithm performs well (Section~\ref{sec:synth-exp}).
We then evaluate how our algorithms perform on a real-world
image classification task (Section~\ref{sec:real-exp}),
and empirically study anti-monotonicity (Section~\ref{sec:exp-assumptions}) and finally
evaluate our algorithms on the real-wold classification task.

%This section is organized as follows: 
%We present our setup in Section~\ref{sec:exp-setup},
%and evaluate the first two goals on synthetic
%datasets in Section~\ref{sec:synth-exp},
%and then on real datasets in Section~\ref{sec:real-exp}.
%We then validate assumptions in Section~\ref{sec:exp-assumptions}.

\subsection{Experimental Setup}\label{sec:exp-setup}

\stitle{Metrics:}
We study multiple metrics in our experiments:
\begin{compactitem}
\item {\em Offline Feature Set Expansion (Metric 1)}: Here, we measure the number of 
feature sets ``expanded'' by our algorithms, which represents the amount of training
required by our algorithm. 
\item {\em Offline Index (Metric 2)}: Here, we measure the total size
of the index necessary to look up the appropriate machine learning
model given a new item and budget constraints.
\item {\em Online Index lookup time (Metric 3)}: Here, we measure the amount of time taken to consult
the index.
\item {\em Online Accuracy (Metric 4)}: Here, we measure the accuracy of the algorithm
on classifying items from the test set.
% \item {\em Feature evaluation time (Metric 5)}: Here, we measure the time to evaluate the 
% feature set that was chosen by the algorithm for the item. 
% \agp{This only makes sense if we ``admit'' that our algorithm
% doesn't really always meet the constraints.}
\end{compactitem}

\noindent 
In the offline case, the reason why we study these two metrics (1 and 2) separately
is because in practice we may be bottlenecked in some cases by the machine learning
algorithm (i.e., the first metric is more important), 
and in some cases by how many machine learning models 
we can store on a parameter server
(i.e., the second metric is more important).
% some cases by the intersection determination algorithm 
% (in which case the second metric is more important); thus
% it is best to separate and independently measure both components.
The reason behind studying the two metrics in the online case is similar.

% The reason why we study these two time components separately
% is because in practice we may be bottlenecked in some cases by the index lookup time
% (in which case the first metric is more important), 
% and some cases by the feature evaluation 
% (in which case the second metric is more important); thus
% it is best to separate and independently measure both components.
%\agp{If index lookup is so important, why would we even
%try to optimize for feature evaluation}

\stitle{Our Algorithms:}
We consider the following algorithms that we have either developed
or adapted from prior work against each other:
\begin{compactitem}
\item \underline\pdom: The optimal algorithm, which requires more storage and precomputation.
\item \underline\gm: The algorithm adapted from Xu {\em et al}~\cite{DBLP:conf/icml/XuWC12}, 
which requires less storage and precomputation
than \pdom, but may expand a sub-optimal set of feature sets that result in lower accuracy
when the item sizes change.
\item \underline\gacc: This algorithm involves a single sequence of \gm (as opposed to multiple 
sequences) prioritizing for the features contributing most to accuracy.
\item \underline\gcost: This algorithm involves a single sequence of \gm prioritizing 
for the features that have least cost.

\end{compactitem}

\stitle{Comparison Points:}
We developed variations of our algorithms to serve as baseline
comparisons for the lattice exploration and indexing portions of an online prediction task:

\begin{compactitem}
\item {\it Lattice Exploration}: \underline{\sc Naive-Expand-All} expands the complete feature set lattice.
\item {\it Indexing}:  While \pdom only indexes points where the dominance relationship
changes, \underline\pdomind \ indexes every intersection point between all pairs of candidate feature sets.
Alternatively, \underline\nlook \ does not create an index and instead scans all candidate
feature sets online.
% \item {\it Online Prediction}: \gacc and \gcost are two extreme variations of \gm that only consider
% accuracy or cost when exploring the feature set lattice.

%\item {\it Metric 1}: We compare against a naive algorithm 
%that expands all feature sets.
%
%\item {\it Metric 2 and 3}:  After expanding a set of feature sets, we compare with two alternative indexes.
%\underline\pdomind indexes every intersection point between all pairs of expanded feature sets
%(in contrast, \pdom only indexes points when the 
%dominance relationship changes).  
%\underline{\sc  Naive-Lookup} scans all candidate feature sets
%and returns the most accurate feature sets whose estimated cost is within the budget.
%
%\item {Metric 4}: We evaluate the estimated 
%compare against \gm, \gacc, and \gcost, which are adaptations of the GreedyMiser~\cite{DBLP:conf/icml/XuWC12} 
%described in Section~\ref{sec:gm-solution}.  

%Unlike \pdom which can be used
%with any machine learning algorithm, this algorithm
%is tightly integrated into a specific machine learning algorithm.
%However, to make the comparison more ``fair'', we modified
%the algorithm such that the algorithm can be applied 
%as a wrapper on top of any machine learning algorithm.
%We describe how we did this in Section~\ref{sec:related}.

\end{compactitem}

\subsection{Synthetic Prediction Experiments}\label{sec:synth-exp}

Our synthetic experiments explore how feature extraction costs, individual feature accuracies,
interactions between feature accuracies, and item size variance affect our algorithms along each of our four metrics.

\stitle{Synthetic Prediction:}
Our first setup uses a synthetic dataset whose item sizes vary
between $1$ and $500$.  To explore the impact of non-constant feature
extraction costs, we use a training set whose sizes are all $1$,
and vary the item sizes in the test dataset.  For the feature sets,
we vary four key parameters:

\begin{compactitem}

\item \noindent{\it Number of features $f_i$}:  We vary the number of features 
in our experiments from 1 to 15.  The default value in our experiments is $12$.

\item \noindent{\it Feature extraction cost $c(f_i,n)$}: We randomly assign the cost
function to ensure a high degree of intersection points.  Each
function is a polynomial of the form $c = a_0 + a_1 n + a_2 n^2$
where the coefficients are picked as follows:  $a_0 \in [0, 100]$,
$a_1 \in [0, \frac{100-a_0}{10}]$, $a_2 \in [0, \frac{100-a_0-a_1}{4}]$.
(The reason why typically $a_0 < a_1 < a_2$ is that $a_1$ is multiplied by $n$,
while $a_2$ is multiplied by $n^2$.)
We expect that typical cost functions are bounded by degree $n^2$
and found that this is consistent with the cost functions from the
real-world task.  Note that \pdom is insensitive to the exact
cost functions, only the intersection points.

\item \noindent{\it Single feature accuracy $a(\{f_i\})$}: 
Each feature's accuracy is sampled to be either {\em helpful} with probability $p$ or 
{\em not helpful} with probability $1-p$. 
If a feature is helpful, then its accuracy is sampled uniformly from within
$[0.7, 0.8]$ and within $[0.5, 0.6]$ if it is not.
% latter
% We assume
% that features are either independent (\underline{INDEP}), correlated
% (\underline{CORR}), or anti-correlated (\underline{ANTI}) to 
% its extraction cost.   \underline{INDEP} categorizes features as
% helpful with probability $p$ and unhelpful with probability $1-p$,
% without regard for the cost function.  The former uniformly samples
% the accuracy within $[0.7, 0.8]$ and within $[0.5, 0.6]$ for the
% latter.  
% \underline{CORR} models a positive linear correlation between the cost and accuracy 
% (that is, the higher the accuracy, the higher the cost of a feature)
% and \underline{ANTI} models
% a negative linear correlation.  
% In both cases, the accuracy linearly varies between $0.5$ and $1$.

\item \noindent{\it Feature interactions $a(F_i)$}: We control how the accuracy
of a feature set $F_i$ depends on the accuracy of its individual
features using a parameterized {\it combiner} function:
$$a_k(F_i) = 1 - \Pi_{f_j \in F_i^k} (1 - a(f_j))$$

\noindent where $F_i^k$ are the top $k$ most accurate features in
$F_i$.  Thus when $k=1$, $F_i$'s accuracy is equal to its most
accurate single feature.  When $k=\infty$, $F_i$'s accuracy increases
as more features are added to the set. 
We will explore $k = 1$ and $\infty$ as two extremes of this 
combiner function.
We denote the combiner function for a specific $k$ value as $cf_{k}$.
Note that for any $k$, the accuracy values are indeed monotone.
We will explore the ramifications
of non-monotone functions in the real-world experiments.
\end{compactitem}

We use the following parameters to specify a specific synthetic
configuration: the number of features $n$, 
%how a feature's accuracy relates to its cost ACC $\in \{$INDEP, CORR, ANTI$\}$ and its parameter
the parameter
$p$, and $k$, the amount the features interact with each other.  In
each run, we use these parameters to generate the specific features,
cost functions and accuracies that are used for all of the algorithms.
Unless specified, the default value of $\alpha$ is $1.2$.
For \gm, the ${\cal L} = \{ 0, 0.001, 0.002, 0.005, 0.01, 0.02, 0.05, 0.1, 0.5, 1, 5, 10 \}$. 
Although we vary ${\cal L}$ over a large range, in practice the majority
give rise to identical sequences because the dominating set of feature sets is fixed for a given item size (as assumed by \gm).

\stitle{Feature Set Expansions (Metric 1):}
We begin by comparing the number of models that \pdom, \gm and \nexp train
as a function of the combiner function and the number of features. 
This is simply the total number of unique feature  
For \pdom and \nexp, the number of feature sets is simply the number
of nodes in the lattice that are expanded,
while for \gm this is simply the number of unique feature sets expanded.

% Since the cost of training a model depends on 
% the specific application, we instead measure the number of feature sets that are expanded (i.e., the number of models that are trained).

\squishframe
\noindent {\bf Metric 1 Summary:}
On the synthetic dataset, for $cf_1$, the number of feature sets (i.e., lattice nodes) 
expanded by \pdom's lattice pruning phase for $\alpha = 1.2$ is at least an order-of-magnitude (10$\times$)
smaller than \nexp, and is similar to \gm.

\begin{figure}[h!]\vspace{-5pt}
\vspace{-5pt}
\centering
\subfigure{\includegraphics[width = 1.5in]{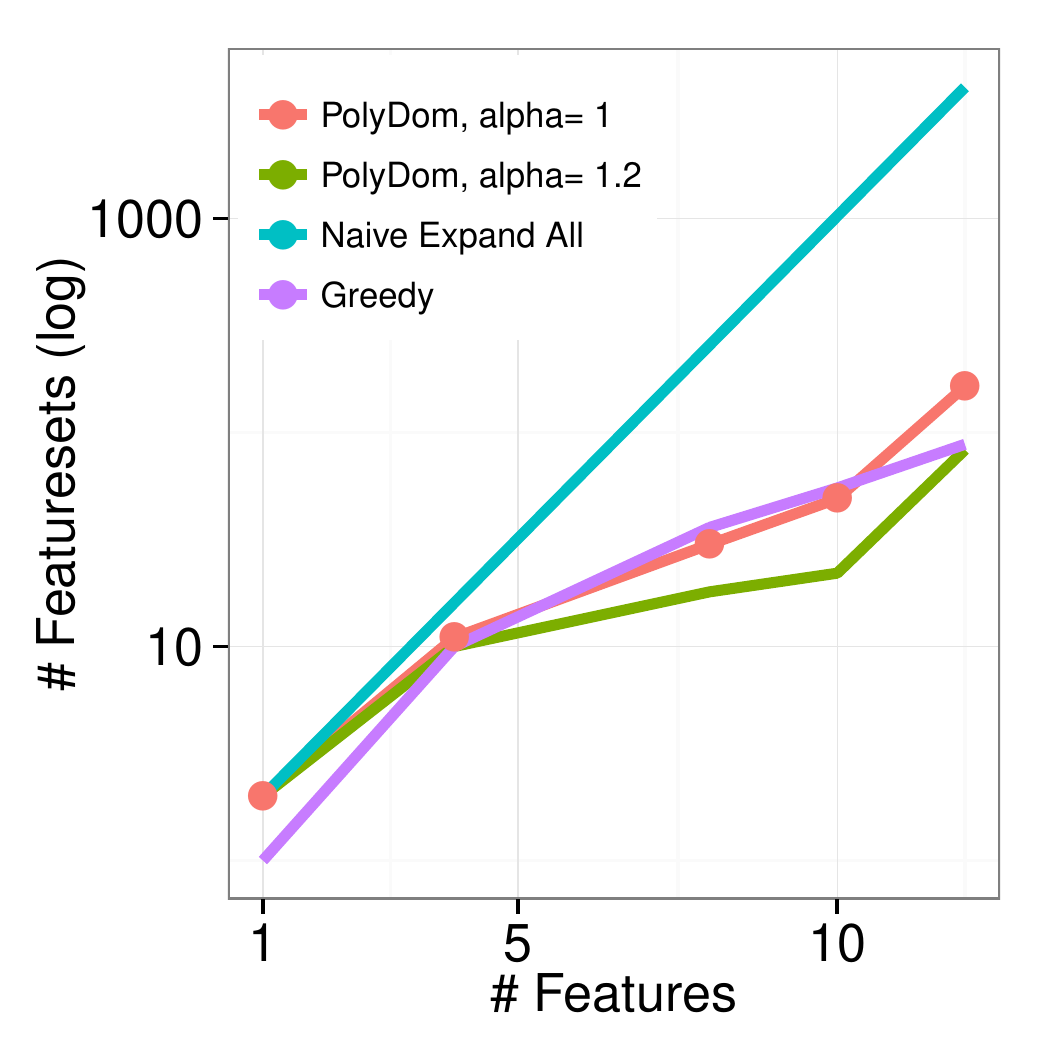}\label{fig:pruning_cf1_synth}}
\subfigure{\includegraphics[width = 1.5in]{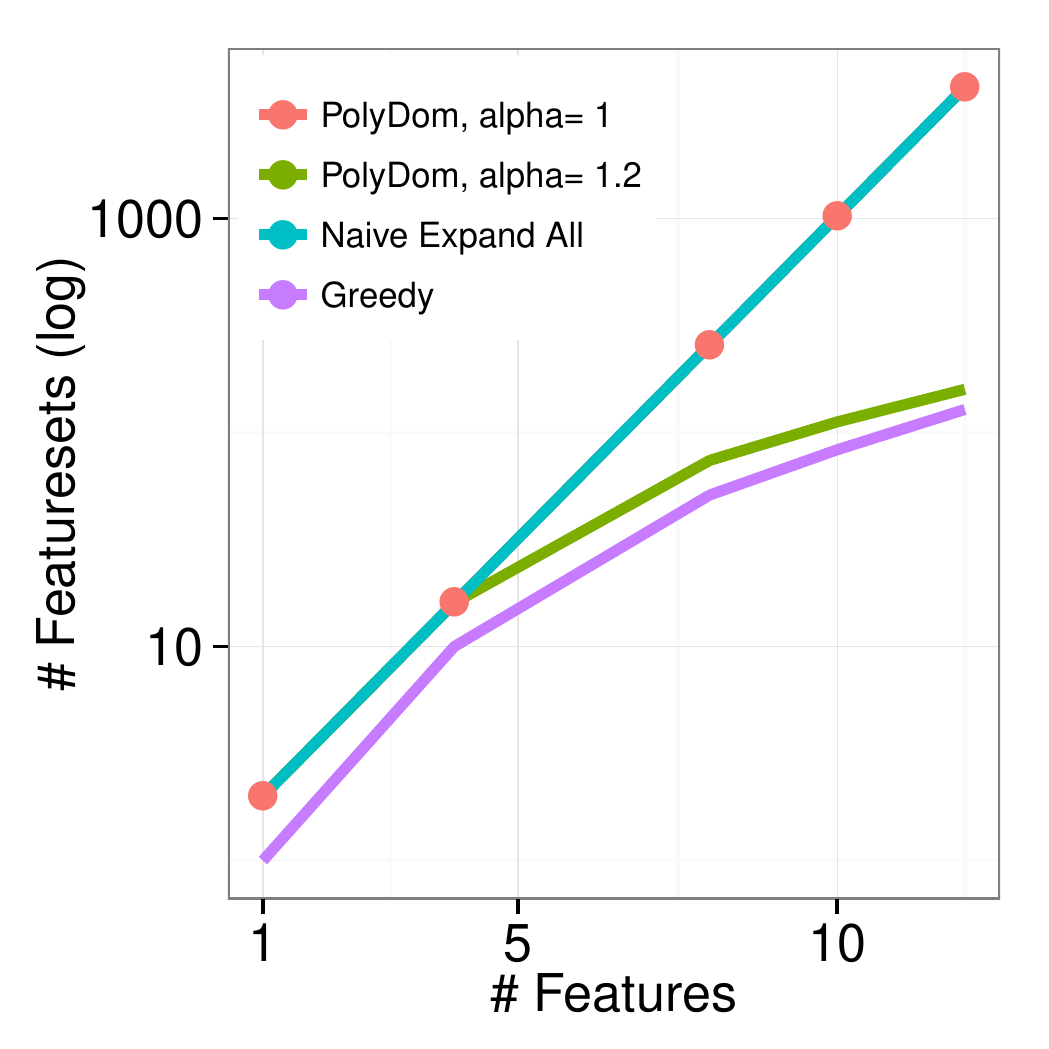}\label{fig:pruning_cf2_synth}}
\vspace{-5pt}\caption{\# of feature sets expanded vs \# features
for (a) $cf_1$ (b) $cf_\infty$}
\vspace{-10pt}\end{figure}

While \pdom with $\alpha = 1$ expands similar to $\alpha = 1.2$ for $cf_1$,
it expands all features for $cf_\infty$.
This is not surprising at all:
$cf_\infty$ is designed to be a scenario where every intermediate feature set
is ``useful'', i.e., none of them get sandwiched between other feature sets.
\frameend
In Figure~\ref{fig:pruning_cf1_synth} and Figure~\ref{fig:pruning_cf2_synth}, 
we depict the number of feature sets
expanded (in log scale) 
as a function of the number of features in the dataset along the $x$ axis,
for $cf_1$ and $cf_\infty$ respectively.
%The plot represents an average of XXX separate runs.
$p$ is set to $0.6$. The plots for other values are similar.

For both combiner functions, 
the total number of possible feature sets 
(depicted as \nexp) scales very rapidly, as expected.
On the other hand, the number of feature sets expanded 
by \pdom for $\alpha = 1.2$ grows at a much slower rate
for both graphs, because \pdom's pruning rules allow it to ``sandwich''
a lot of potential feature sets and avoid expanding them.
Consider first combiner function
(i.e., Figure~\ref{fig:pruning_cf1_synth})
For 10 features, \nexp
expands around 1000 feature sets, 
while \pdom for $\alpha = 1$ expands about 50,
and $\alpha = 1.2$ expands about 30.
% The rate of increase for
% $p = 0.6$ is higher than that for $p = 1$;
% this is not surprising because a $p$ value
% closer to $0.5$ ensures that the individual
% feature accuracies are more ``varied'',
% and hence the number of expanded feature sets
% is higher.
% This holds true even for combiner function $2$,
% although the difference between the two $p$ values
% is not as significant.
We find that the ability to effectively prune
the lattice of feature sets depends quite a bit on the combiner function.
While \pdom with $\alpha = 1.2$ continues to perform similarly.
\pdom $\alpha = 1$ expands as much as \nexp;
this is not surprising given that all intermediate feature sets
have accuracy values strictly greater than their immediate children.
In comparison, \gm expands about as many features as \pdom with 
$\alpha = 1.2$ but with a slower growth rate
as can be seen from both $cf_1$ and $cf_\infty$;
this is not surprising because in the worst case \gm
expands $|{\cal L}| \times |{\cal F}|^2$.

% Further, we see that the ability to effectively prune the lattice depends heavily on the nature of
% the combiner function: the first combiner function, where the set of singleton features dominate
% the rest of the feature sets requires far fewer feature sets to be expanded.

\stitle{Indexing Size and Retrieval (Metric 2 and 3):} Here, we measure the indexing
and retrieval time of the \pdom algorithms,
which use a more complex indexing scheme
than the \gm algorithms.

\squishframe
\noindent {\bf Metric 2 and 3 Summary:} On the synthetic dataset, 
especially for larger numbers of features, 
the size of the \pdom index is significantly smaller than the size
of the \pdomind index, and almost as small as the \nlook index.
However, while \nlook has a smaller index size, \nlook's retrieval time is much larger
than \pdom (for multiple values of $\alpha$), making it an unsuitable candidate.
% this reduction in size does not come at the cost of index lookup time:
% the retrieval time performance of the poly-dom index is similar to but slightly better
% than that of \pdomind, and significantly better than \nlook.
\frameend

\begin{figure}[h!]\vspace{-5pt}
\vspace{-5pt}
\centering
\subfigure{\includegraphics[width = 1.5in]{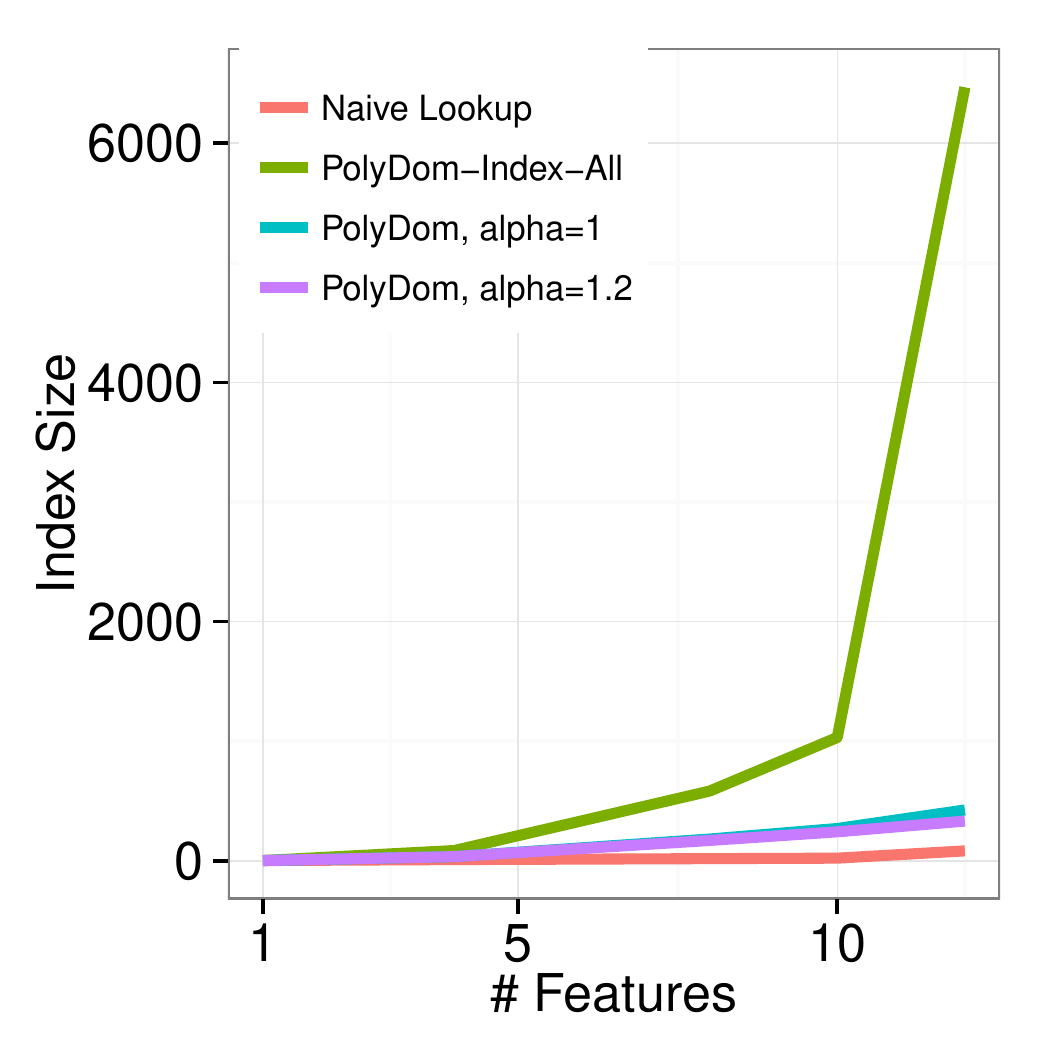}\label{fig:index_cf1_synth}}
\subfigure{\includegraphics[width = 1.5in]{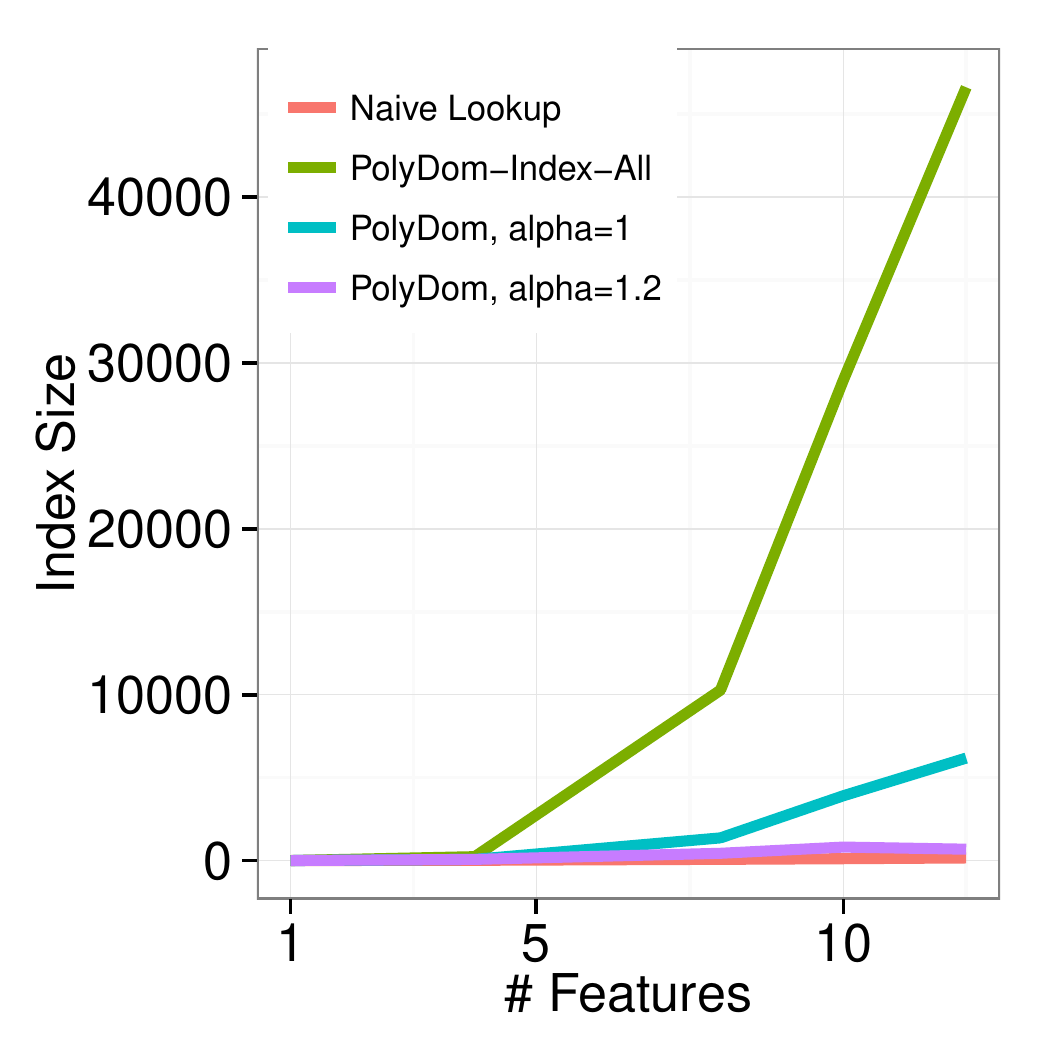}\label{fig:index_cf2_synth}}
\vspace{-5pt}\caption{Index Size for (a) $cf_1$ (b) $cf_\infty$}
\vspace{-10pt}\end{figure}

\noindent In Figure~\ref{fig:index_cf1_synth} and Figure~\ref{fig:index_cf2_synth}, we plot,
for the two combiner functions the 
total size of the poly-dom index as the number of features is increased
(for $\alpha = 1.2, p = 0.6$.)
Consider the case when the number of features is $10$ for $cf_1$:
here, \pdom and \nlook's index size are both less than 200, 
\pdomind's index size is at the 1000 mark, and rapidly increases to 6000 for 12 features,
making it an unsuitable candidate for large numbers of features.
The reason why \pdom's index size is smaller than \pdomind is because \pdom
only indexes those points where the dominance relationship changes, while
\pdomind indexes all intersection points between candidate feature sets.
\nlook, on the other hand, for both $cf_1$ and $cf_\infty$ only needs to record
the set of candidate sets, and therefore grows slowly as well.

% \nlook's index is much larger than the two \pdom-based strategies because
% the number of candidate sets grows very rapidly as the number of features grows,
% even if those candidate sets may very well be dominated between [0, maximum size]:
% consider two cost polynomials: $n+n^2$, and $100000\times n$, where the maximum size
% is $100$; here, $n + n^2$ can never dominate $100000\times n$ between $0$ and $100$.
% Even so, \nlook would record both the candidate sets as part of the index,
% while \pdom will only record the intersection points where dominance changes.
% The behavior for the other combiner function is similar, except that
% the \nlook algorithm's index size grows even faster from the start. 

% The anomalous behavior for \nlook until the point where the number of features is
% $5$ is probably because we only record two candidate sets until that point, even though the number 
% of ....
% \agp{Maybe we use mean to get rid of this anomaly?}

\begin{figure}[h!]\vspace{-5pt}
\vspace{-5pt}
\centering
\subfigure{\includegraphics[width = 1.5in]{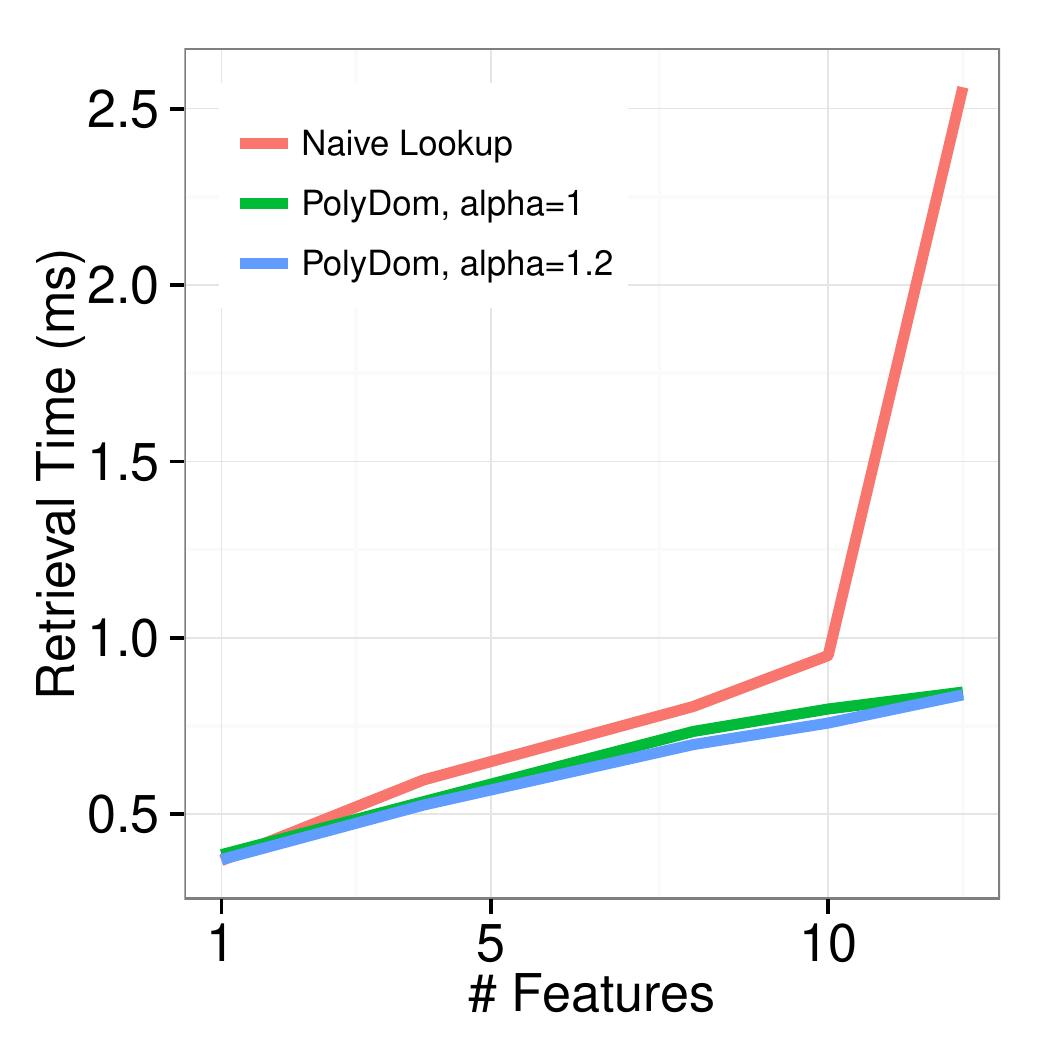}\label{fig:test_cf1_synth}}
\subfigure{\includegraphics[width = 1.5in]{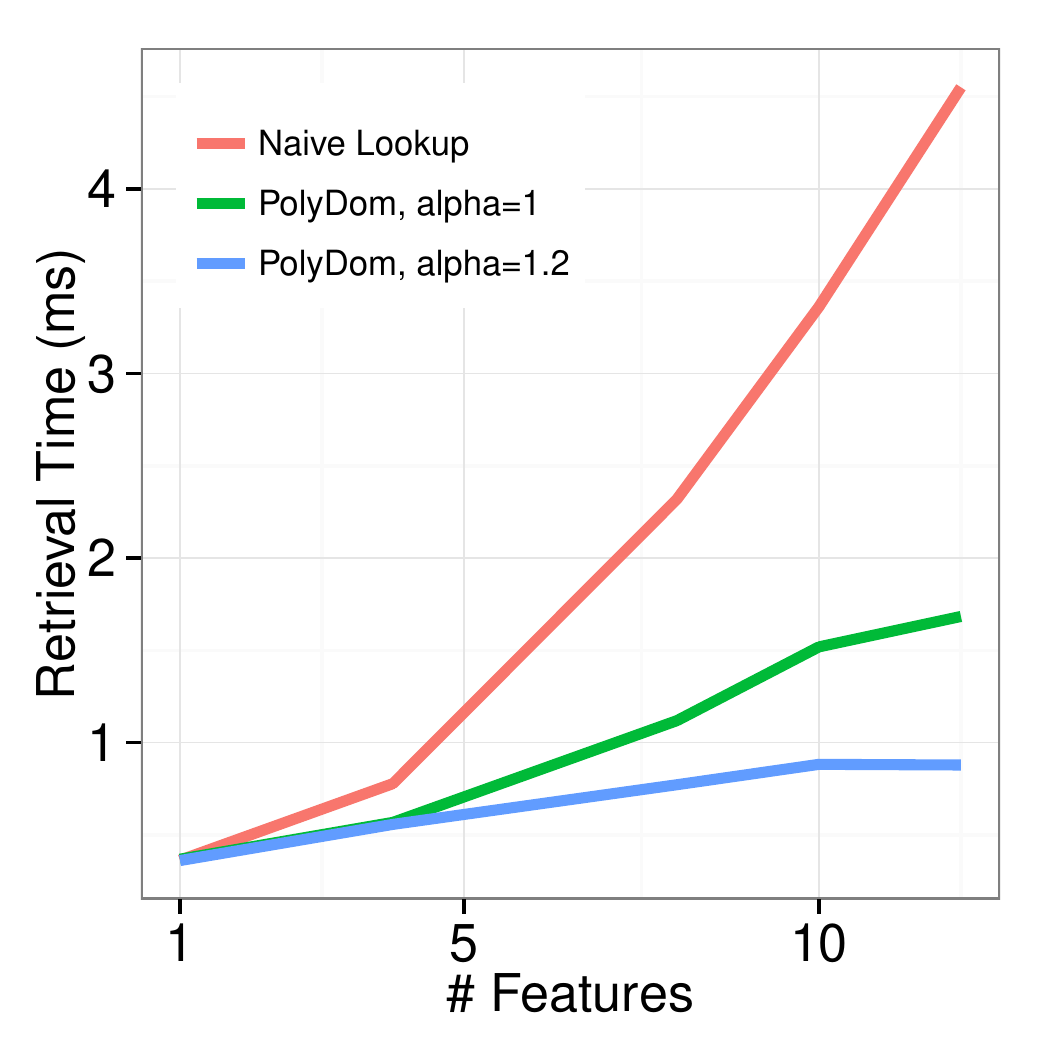}\label{fig:test_cf2_synth}}
\vspace{-5pt}\caption{retrieval time for (a) $cf_1$ (b) $cf_\infty$}
\vspace{-10pt}\end{figure}

On the other hand, for retrieval time,
depicted in Figures~\ref{fig:test_cf1_synth} and \ref{fig:test_cf2_synth}, 
we find that the \pdom's clever indexing scheme
does much better than \nlook, since we
have organized the feature sets in such a way
that it is quick to retrieve the appropriate feature set given
a cost budget
On the other hand, \nlook does significantly worse than \pdom, 
since it linearly scans
all candidate feature sets to pick the best one ---
especially as the number of features increases.


\begin{figure}[h!]\vspace{-5pt}
\vspace{-5pt}
\centering
\subfigure{\includegraphics[width = 1.5in]{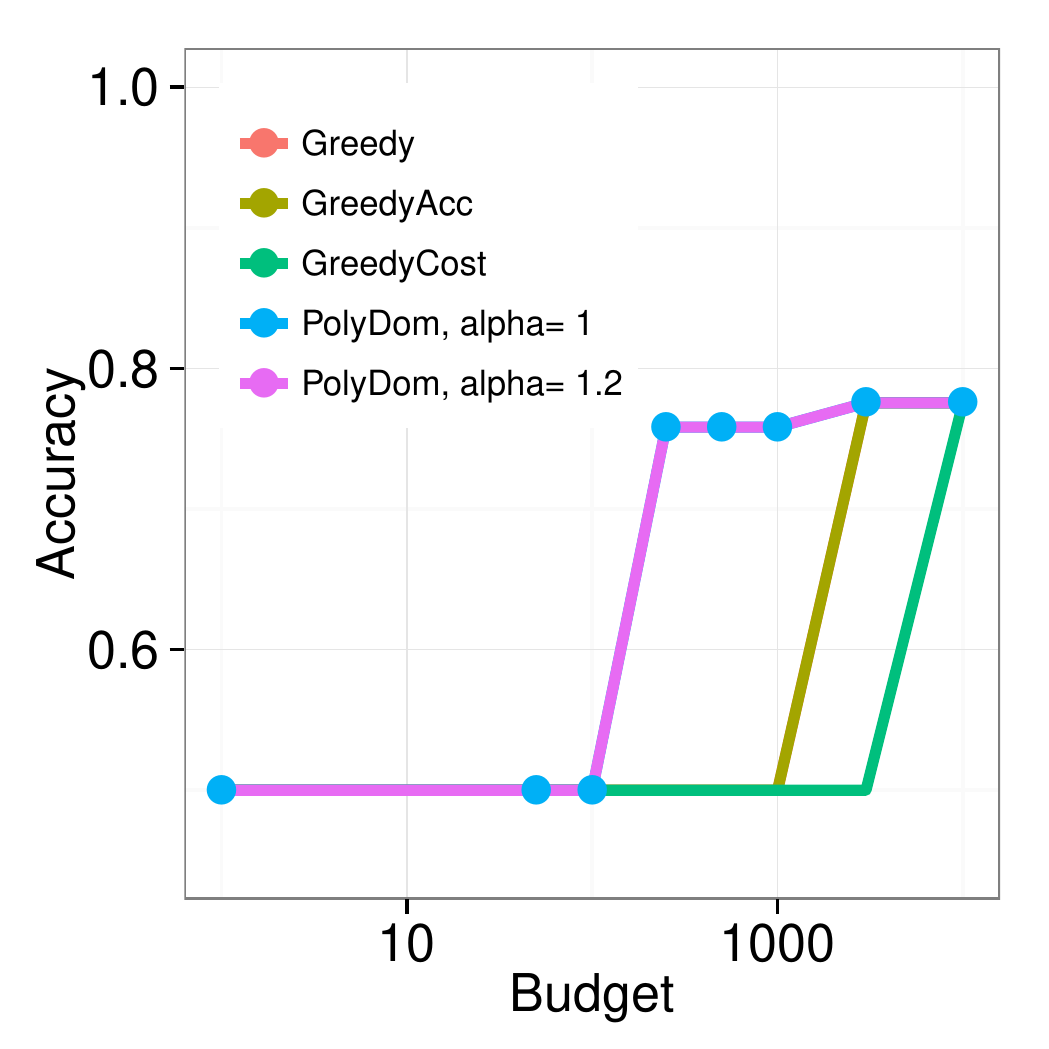}\label{fig:estacc_vary_budget_cf1_synth}}
\subfigure{\includegraphics[width = 1.5in]{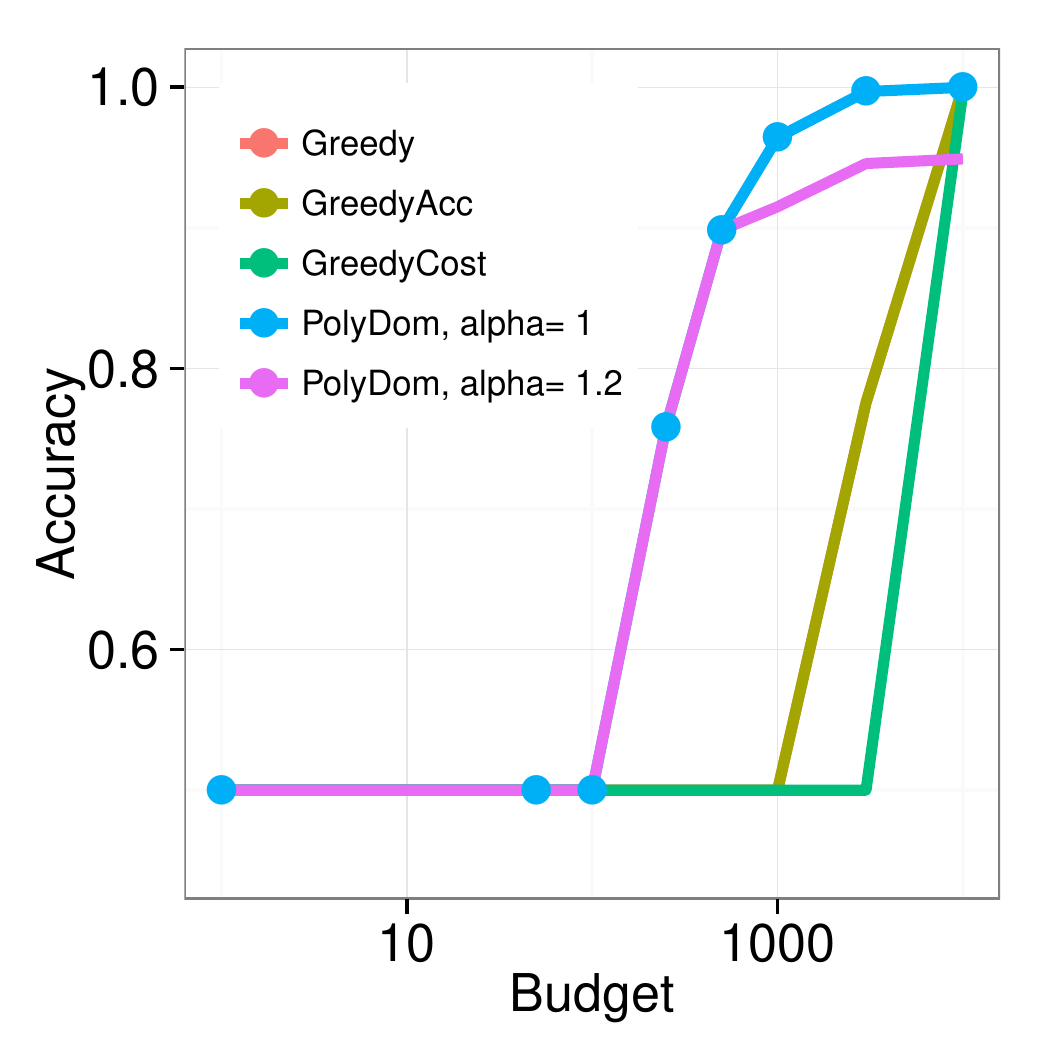}\label{fig:estacc_vary_budget_cf2_synth}}
\vspace{-5pt}\caption{Accuracy as a function of budget (a) $cf_1$ (b) $cf_\infty$}
\vspace{-10pt}\end{figure}

\stitle{Real-Time Accuracy (Metric 4):} We now test the accuracy of the eventual model
recommended by our algorithm.

\squishframe
\noindent {\bf Metric 4 Summary:} On synthetic datasets, for $cf_1$ and $cf_\infty$, 
over a range of budgets, \pdom (with both $\alpha = 1$ or $1.3$), returns
models with greater accuracies than \gm and \gacc, which returns models with greater
accuracies than \gcost.
Often the accuracy difference (for certain budgets) between \pdom and \gm,
or between \gm and \gcost can be as high as 20\%.
% our algorithm does much better than \gm,
% with an accuracy increase of XX\% on average for $\alpha = X$,
% and YY\% on average for $\alpha = Y$.
\frameend
In figure~\ref{fig:estacc_vary_budget_cf1_synth} and \ref{fig:estacc_vary_budget_cf2_synth},
we plot the accuracy as a function of budget
for \pdom with $\alpha = 1$ and $1.2$, and for \gm, \gacc and \gcost.
For space constraints, we fix the item size to $50$ and use $12$ features.
$\alpha = 1$ and $1.2$ is almost always
better than \gm.
For instance, consider budget 1000 for $cf_1$
\pdom with $\alpha = 1\&1.2$ has an accuracy of about 80\%, while
\gm, \gacc and \gcost all have an accuracy of about 50\%; 
as another example, consider budget 1000 for $cf_{\infty}$,
where \pdom with $\alpha = 1\&1.2$ has an accuracy of more than 90\%,
while \gm, \gacc and \gcost all have accuracies of about 50\%.
In this particular case, this may be because \gm, \gacc, and \gcost 
all explore small portions of the lattice and may get stuck in local optima.
That said, apart from ``glitches'' in the mid-tier budget range, all algorithms
achieve optimality for the large budgets, and are no better than random for the low budgets.

Further, as can be seen in the figure \gm does better than \gcost, and similar to \gacc.
We have in fact also seen other instances where \gm does better than \gacc, and similar to \gcost.
Often, the performance of \gm is similar to one of \gacc or \gcost.
% This seems to indicate that setting ${\cal L} = \{-\infty, \infty\}$
% may be sufficient to give as good a performance for \gm without any tuning.

% \agp{figure\ref{fig:estacc_vary_budget_cf2_synth} fixes: accuracy instead of acc, and we need more datapoints between 1e+02 and 1e+03, that's where all the interesting stuff is. We can cap off at 1e+05 if we want}

\begin{figure}[h!]\vspace{-5pt}
\vspace{-5pt}
\centering
\subfigure{\includegraphics[width = 1.5in]{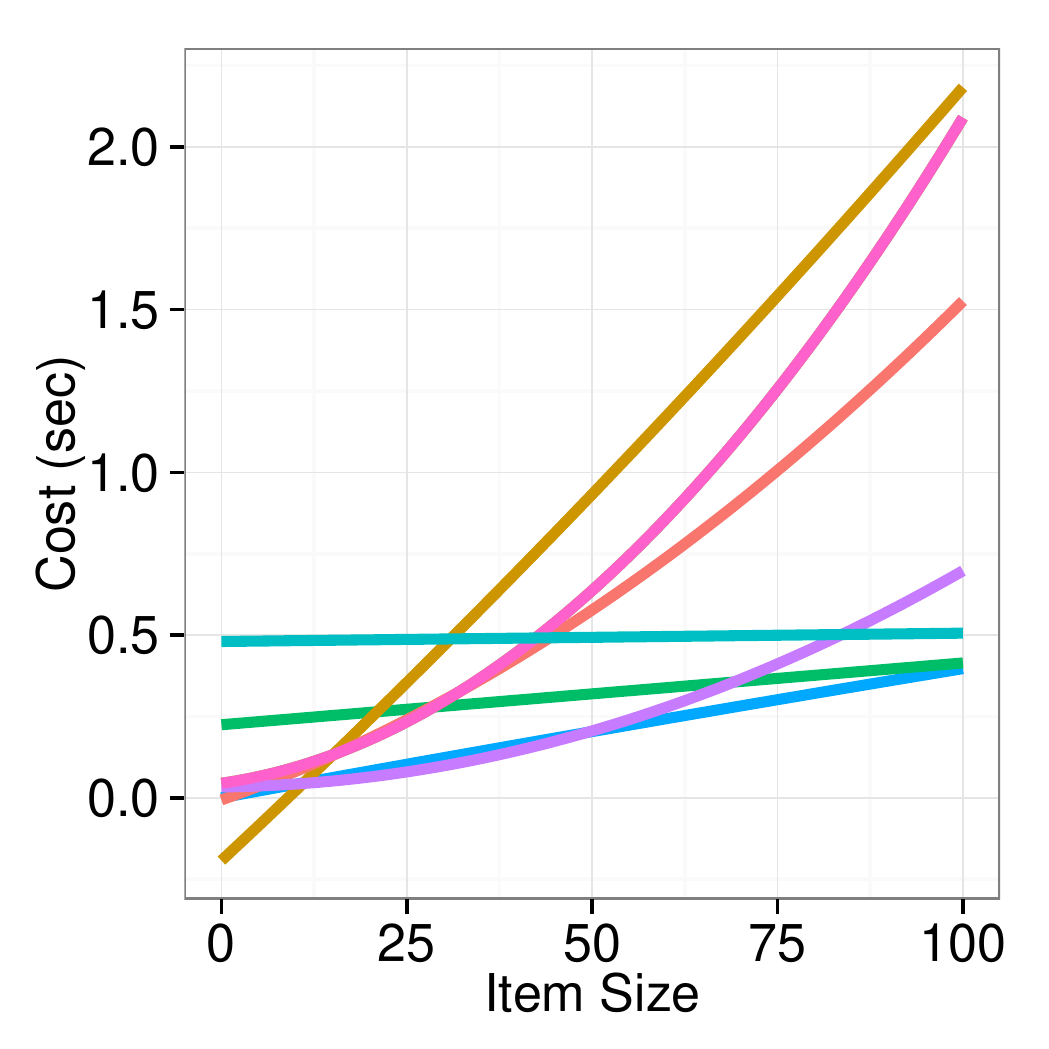}\label{fig:cost_vary_size}}
\subfigure{\includegraphics[width = 1.5in]{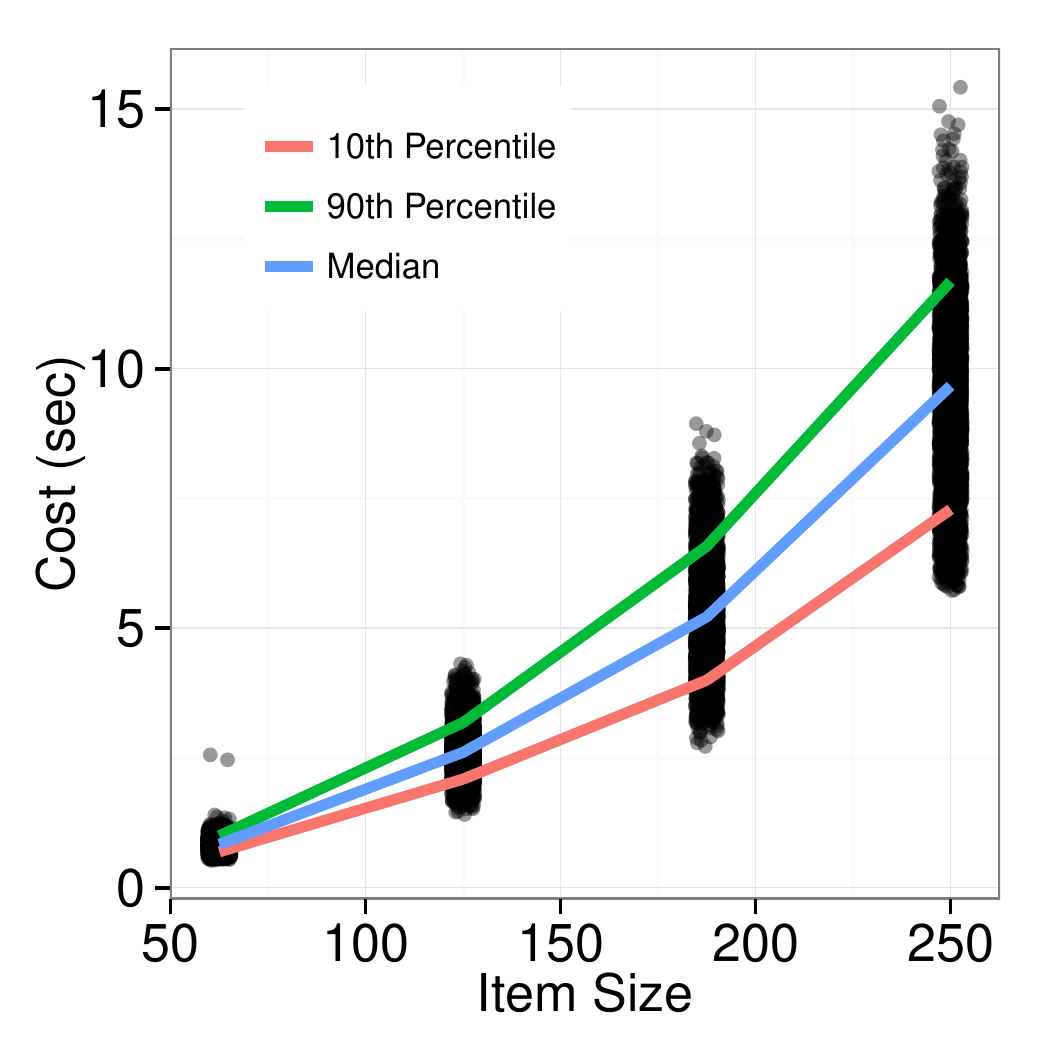}\label{fig:cost_vary_real}}
\vspace{-5pt}\caption{Varying size for (a) many features (b) one feature, but plotting quantiles}
\vspace{-10pt}\end{figure}

\begin{figure*}[ht!]
\vspace{-5pt}
\centering
\subfigure{\includegraphics[width = 1.5in]{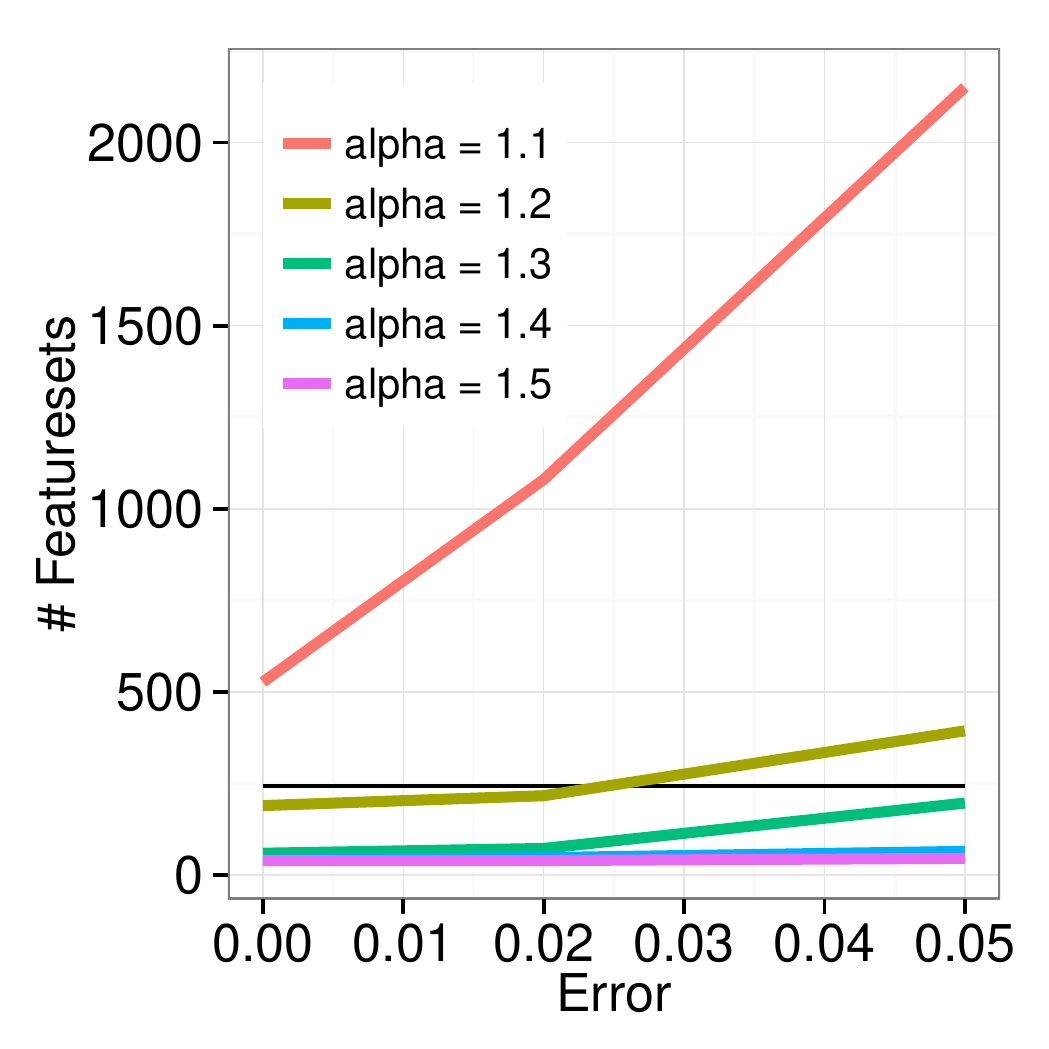}\label{fig:pruning_real}}
\subfigure{\includegraphics[width = 1.5in]{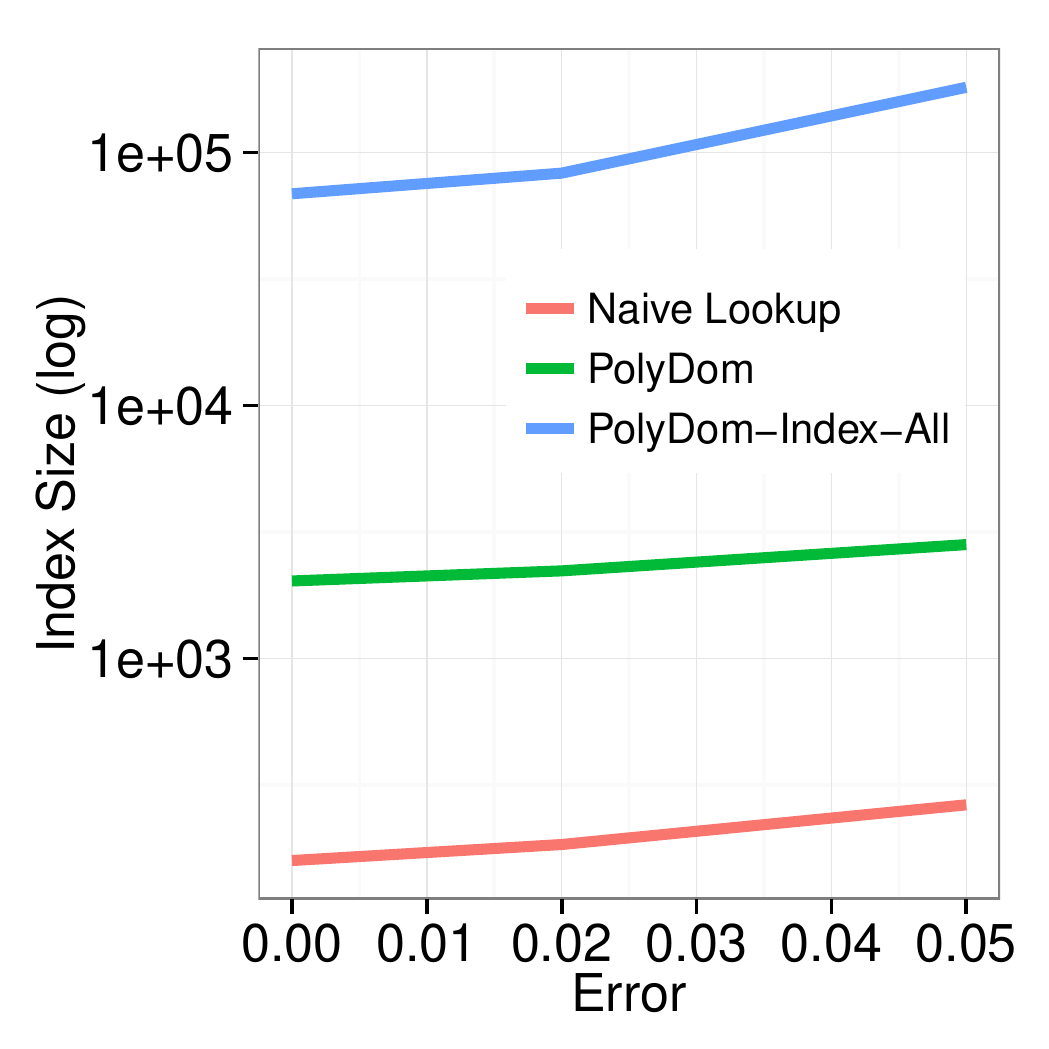}\label{fig:index_vary_e_real}}
\subfigure{\includegraphics[width = 1.5in]{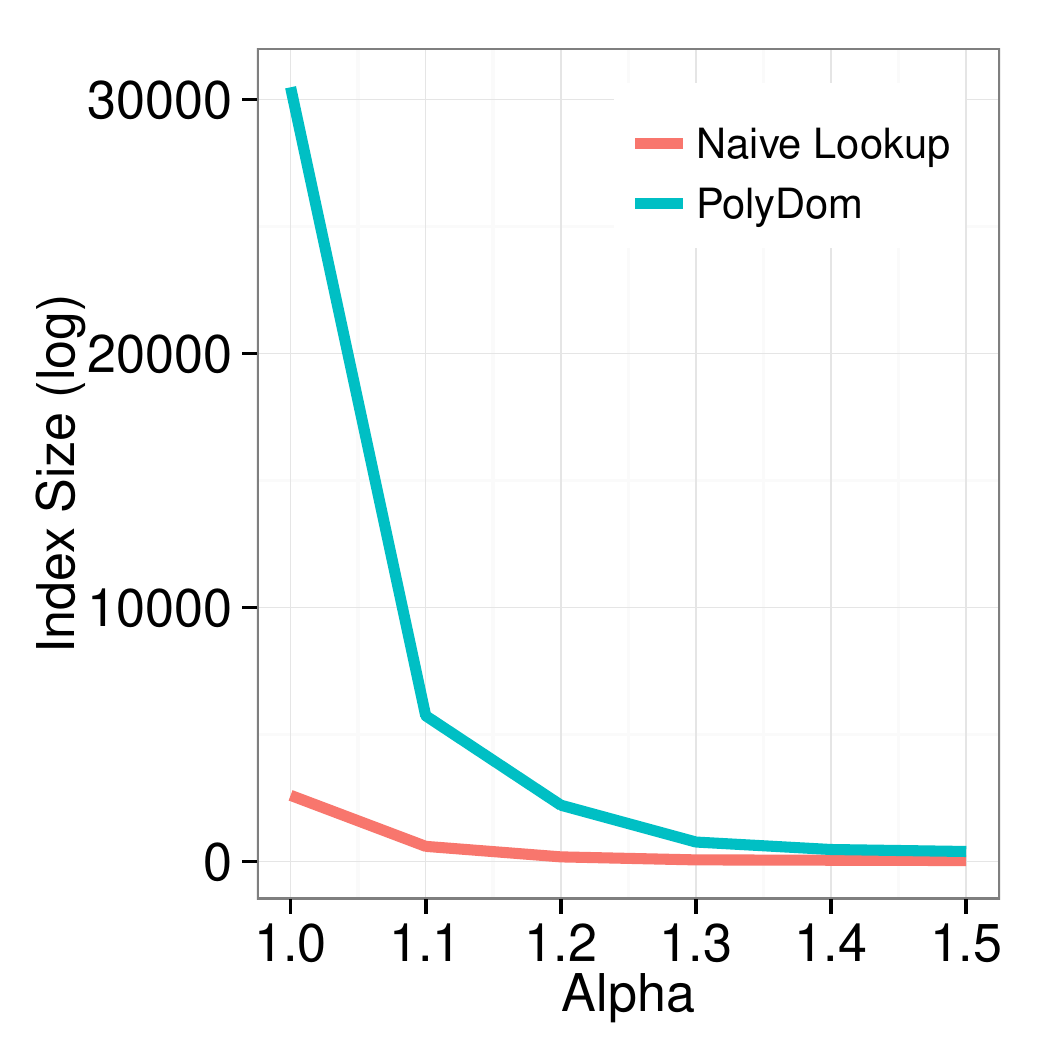}\label{fig:index_vary_alpha_real}}
\subfigure{\includegraphics[width = 1.5in]{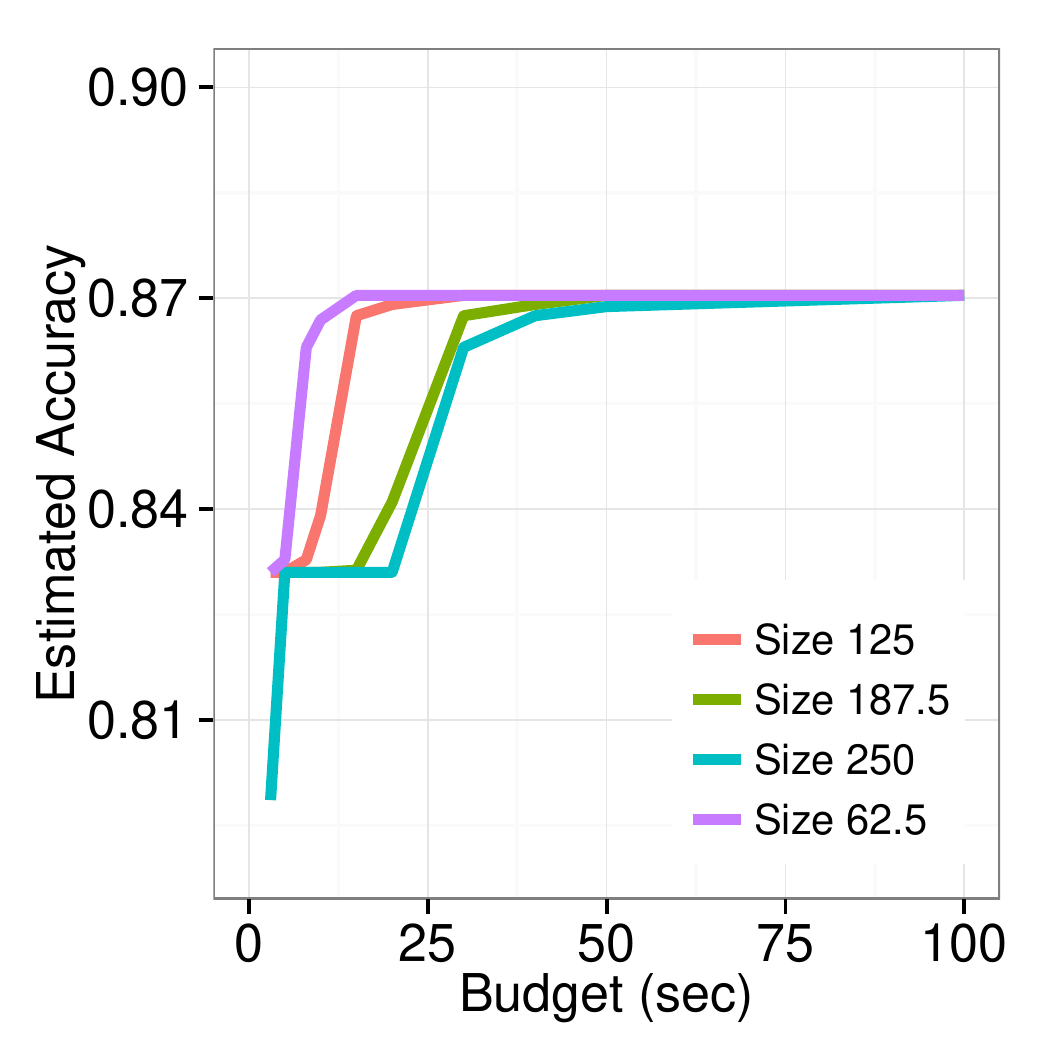}\label{fig:estacc_vary_budget_real}}
\vspace{-5pt}\caption{(a) Feature sets expanded (b) Index Size on varying $e$ (c) Index size on varying $\alpha$ (d) Accuracy vs. Budget}
\vspace{-10pt}
\end{figure*}

\subsection{Real Dataset Experiments}\label{sec:real-exp}

\stitle{Real Dataset:}
This subsection describes our experiments using a real image classification dataset~\cite{image-dataset}.
The experiment is a multi-classification task to identify each image as one
out of 15 possible scenes.  There are 4485 labeled $250\times 250$ pixel images in the original dataset.  To  test
the how our algorithms perform on varying image sizes, we 
rescale them to $65\times 65$, $125\times 125$ and $187\times 187$ pixel sizes.  Thus in total, our dataset contains 17900 images.
We use 8000 images as training and the rest as test images.

The task uses 13 image classification features (e.g., SIFT and GIST features) 
that  vary significantly in cost and accuracy.
In Figure~\ref{fig:cost_vary_size}, we plot the cost functions of
eight representative features as a function of $n$, the item size,
that we have learned using least squares curve fitting to the median cost at each training image size.
As can be seen in the figure, there are some features
whose cost functions are relatively flat (e.g., gist),
while there are others that are increasing linearly (e.g., geo\_map8$\times$8)
and super-linearly (e.g., texton).

However, note that due to variance in feature evaluation time,
we may have cases where the real evaluation cost does not exactly match
the predicted or expected cost. 
In Figure~\ref{fig:cost_vary_size}, 
we depict the 10\% and 90\% percentile of the cost given
the item size for a single feature.
As can be seen in the figure, there is significant variance ---
especially on larger image sizes.

To compensate for this variation, we compute the cost functions using the worst-case
extraction costs rather than the median.  
In this way, we ensure that the predicted models in the experiment are always within budget.
Note that \gm does not need to do this 
since it can seamlessly scale up/down
the number of features evaluated as it traverses
the sequence corresponding to a given $\lambda$.  We did not consider this dynamic approach for
the \pdom algorithm.

% \agp{earlier text: for each candidate feature set that we eventually use, 
% we expand a sequence of feature sets leading all the way to
% the empty set such that we can always terminate 
% before the cost constraint. 
% As we will see later on,
% this leads to a modest increase in the number of feature sets
% expanded.} 

The ``black box'' machine learning algorithm we use is a Linear
classifier using stochastic gradient descent learning with hinge
loss and L1 penalty.  We first train the model over the training
images for all possible combinations of features and cache the
resulting models and cross-validation (i.e., estimated) accuracies.
The rest of the experiments can look up the cached models rather than re-train 
the models for each execution.

For this dataset, our default values for $\alpha, e$ are $1.2, 0$, respectively
As we will see in the following, the impact of $e$ is small, even though
our experiments described appendix~\ref{sec:exp-assumptions} find that $e \ge 0$.

\stitle{Feature Set Expansions (Metric 1):} 
\squishframe
\noindent {\bf Metric 1 Summary:}
On the real dataset, the number of feature sets 
expanded by \pdom's offline lattice pruning phase for $\alpha = 1.2$ is $20\times$ 
smaller than \nexp, with the order of magnitude increasing as $\alpha$ increases,
and as $e$ decreases. 
\gm expands a similar number of feature sets as \pdom with $\alpha = 1.2$.
\frameend
In Figure~\ref{fig:pruning_real}, 
we depict the number of feature sets
expanded (in log scale) 
as a function of the tolerance to non-monotonicity 
$e$ along the $x$ axis, for \pdom with values of $\alpha = 1.1, \ldots, 1.5$,
and for \gm.

As can be seen in the figure, 
while the 
the total number of possible feature sets is close to $8200$
(which is what \nexp would expand), 
the number of feature sets by \pdom is always
less than $400$ for $\alpha = 1.2$ or greater, and
is even smaller for larger $\alpha$s
(the more relaxed variant induces
fewer feature set expansions).
\gm (depicted as a flat black line)
expands a similar number of feature sets as
\pdom with $\alpha = 1.2$.
On the other hand, for $\alpha = 1.1$, 
more than 1/4th of the feature sets are expanded.

The number of feature sets expanded also
increases as $e$ increases
(assuming violations of monotonicity
are more frequent leads
to more feature set expansions).
% However, even with $\alpha = 1$, the number of feature sets expanded is a small fraction
% of the total number of feature sets. \agp{THIS WOULD BE A GOOD STATEMENT TO MAKE}

\begin{figure}[h!]\vspace{-5pt}
\vspace{-5pt}
\centering
\subfigure{\includegraphics[width = 1.5in]{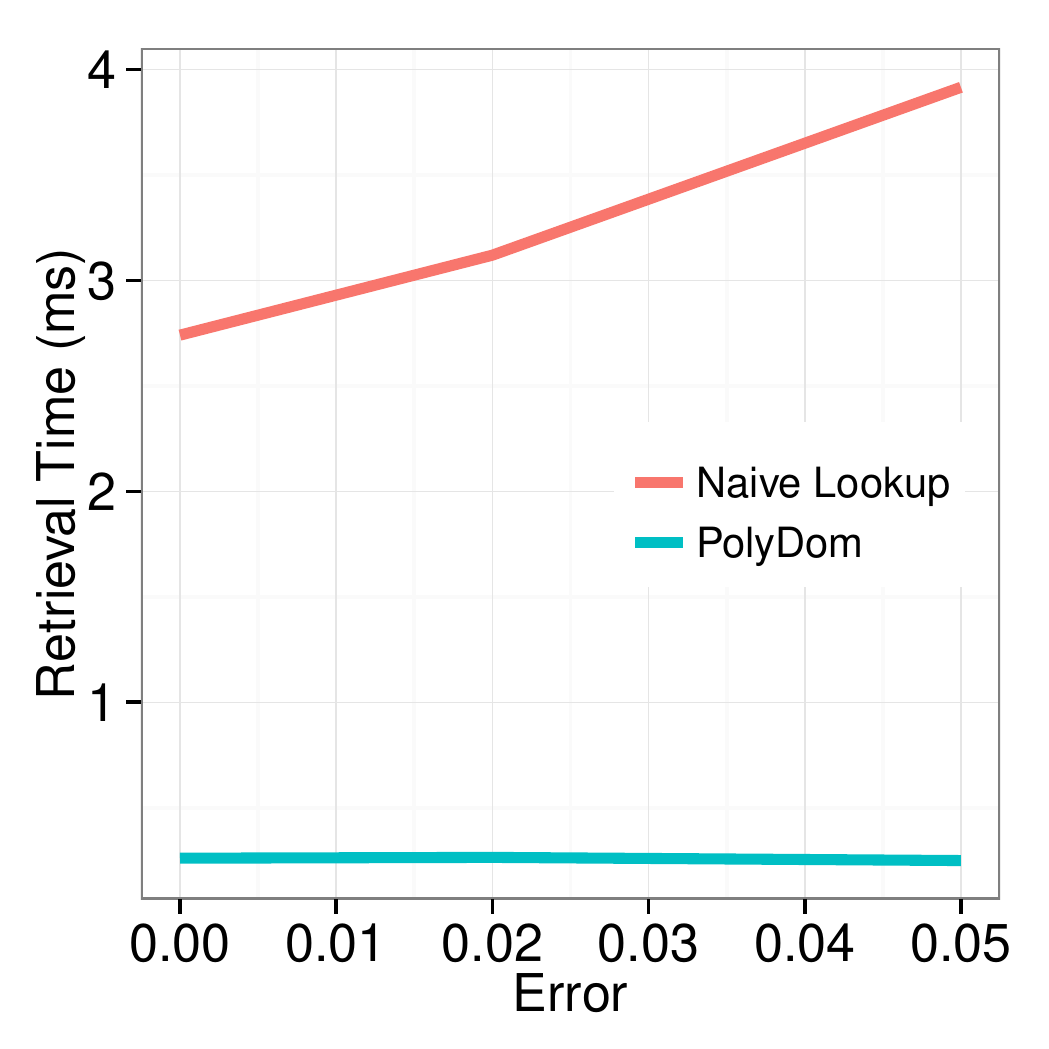}\label{fig:test_vary_e_real}}
\subfigure{\includegraphics[width = 1.5in]{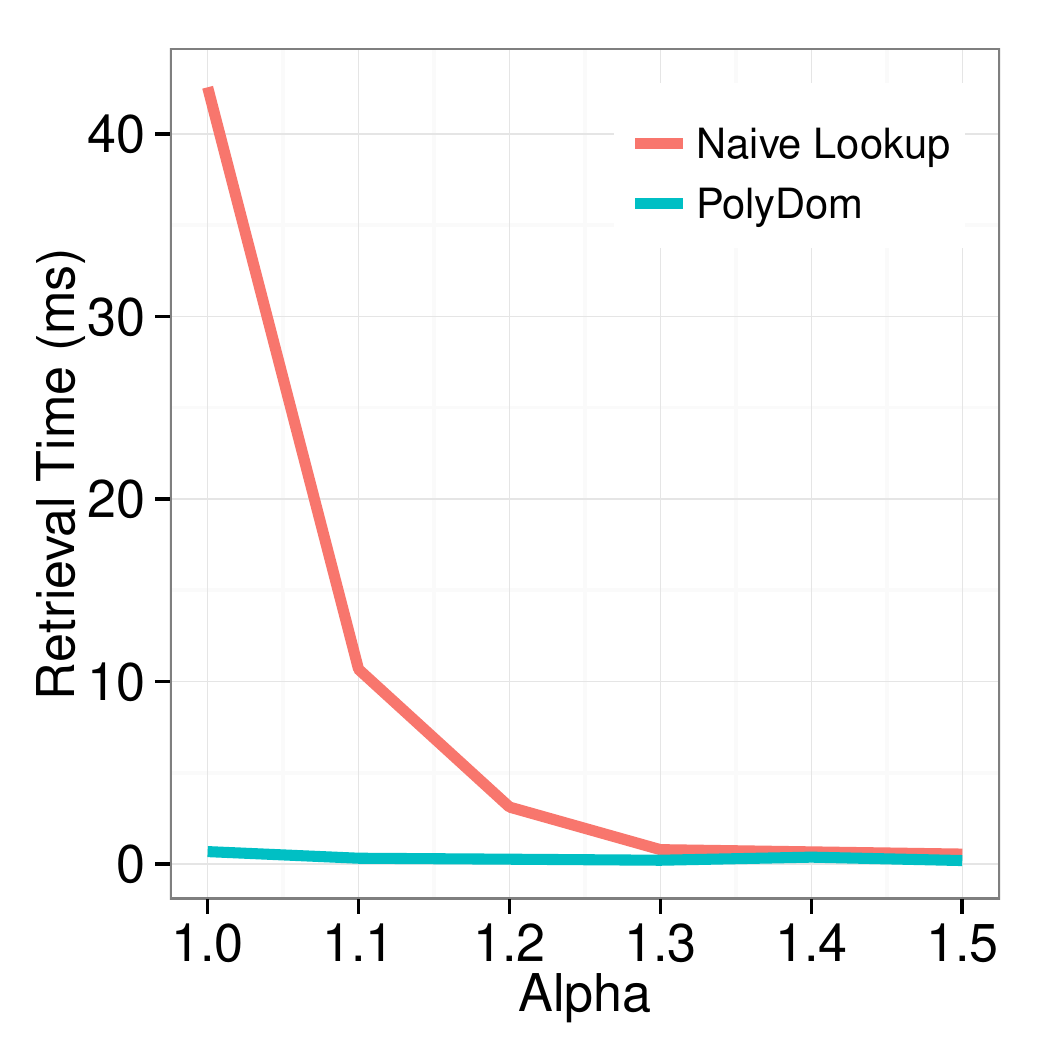}\label{fig:test_vary_alpha_real}}
\vspace{-5pt}\caption{Retrieval time on varying (a) $e$ (b) $\alpha$}
\vspace{-5pt}\end{figure}

\stitle{Indexing Size and Retrieval (Metric 2 and 3):} 
\squishframe
\noindent {\bf Metric 2 and 3 Summary:} On the real dataset the size of the index
for \pdom is two orders of magnitude smaller than \pdomind, while \nlook
is one order of magnitude smaller than that.
The index size increases as $e$ increases and decreases as $\alpha$ increases.
However, the retrieval time for \pdom is minuscule compared to the retrieval time
for \nlook.
\frameend
In Figure~\ref{fig:index_vary_e_real}, we plot the 
total index size as the tolerance to non-monotonicity is increased
(for $\alpha = 1.3$.)
As can be seen in the figure, the index size for \pdom grows slowly
as compared to \pdomind, while \nlook grows even slower.
Then, in Figure~\ref{fig:index_vary_alpha_real}, we display
the total index size that 
decreases rapidly as $\alpha$ is increased.
% \pdom is not better than \nlook for large $\alpha$.
% However, for large $\alpha$ the index size is anyway small,
% so having a larger index size than \nlook is not problematic.

On the other hand, if we look at retrieval time,
depicted in Figures~\ref{fig:test_vary_e_real} and \ref{fig:test_vary_alpha_real} 
(on varying $e$ and on varying $\alpha$ respectively), 
we find that \nlook is much worse than \pdom --- \pdom's indexes
lead to near-zero retrieval times, while \nlook's retrieval time is significant,
at least an order of magnitude larger.
Overall, we find that as the number of candidate sets under consideration grows
(i.e., as $\alpha$ decreases, or $e$ increases), we find
that \pdom does much better relative to \pdomind in terms of space considerations,
and does much better relative to \nlook in terms of time considerations.
This is not surprising: the clever indexing scheme used by \pdom pays richer dividends
when the number of candidate sets is large.

\begin{figure}[h!]\vspace{-5pt}
\centering
\subfigure{\includegraphics[width = 2.5in]{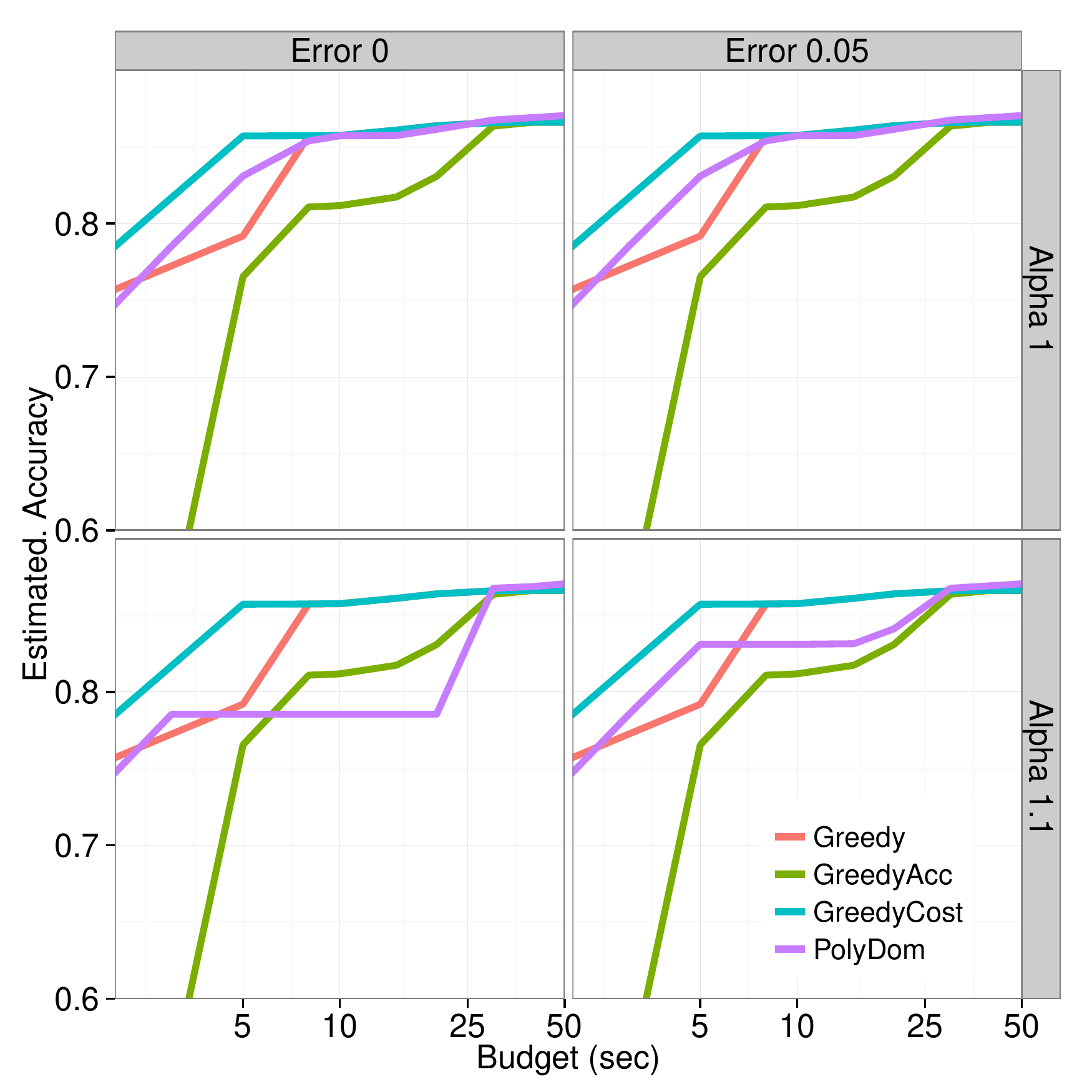}\label{fig:estacc_heatmap_real}}
\vspace{-5pt}\caption{Estimated accuracy vs budget on varying
$\alpha$ and $e$}
\vspace{-10pt}\end{figure}

\stitle{Real-Time Accuracy (Metric 4):} 
\squishframe
\noindent {\bf Metric 4 Summary:} On the real dataset,
we find that while \pdom still performs well compared to 
other algorithms for small $\alpha$, some \gm-based algorithms are competitive,
and in a few cases, somewhat surprisingly, better than \pdom.
All other things being fixed, the accuracy increases
as the item size decreases, the budget increases, the $\alpha$ decreases,
and $e$ increases.
\frameend
\noindent In Figure~\ref{fig:estacc_heatmap_real}, we plot the 
average estimated
accuracy of the model retrieved given a budget for various values
of $\alpha$ and $e$ for 
\pdom, \gm, \gacc, and \gcost across a range of image sizes. 
As can be seen in the figure, 
\pdom dominates \gm and \gacc apart from the case when $\alpha = 1.1$.
Even for $\alpha = 1.1$, \pdom dominates \gacc when $e = 0.05$:
here, we see that a higher $e$ leads to better performance,
which we did not see in other cases.

Perhaps the most surprising aspect of this dataset is that
\gcost dominates all the others overall. 
While the reader may be surprised that a \gm-based algorithm can outperform 
\pdom with $\alpha = 1$, recall that the \gm algorithms are any-time algorithms
that can adapt to high variance in the feature evaluation cost, as opposed to
\pdom, which provisions for the worst-case and does not adapt to the variance.
In future work, we plan to explore any-time variants of \pdom, 
or hybrid variants of \pdom with \gm-based
algorithms.

In Figure~\ref{fig:estacc_vary_budget_real}, we plot the
estimated accuracy of the model retrieved as a function
of the budget for various image sizes (across
a number of images of the same size).
As can be seen in the figure, the estimated
accuracy is higher for the same size of image,
as budget increases.
Also, the estimated accuracy is higher for the same budget,
as image size decreases (as the image size decreases,
the same budget allows us to evaluate more features
and use a more powerful model.)

\section{Related Work}\label{sec:related}

Despite its importance in applications, 
cost-sensitive real-time classification is not a particularly well-studied problem:
typically, a feature selection algorithm~\cite{saeys2007} 
is used to identify a set of inexpensive features that are used with an 
inexpensive machine learning model (applied to all items, large or small), 
and there are no dynamic decisions enabling us to use a
more expensive set of features if the input parameters allow it. 
This approach ends up giving us a classifier that is sub-optimal
given the problem parameters.
This approach has been used for real-time classification 
in a variety of scenarios, including:
sensor-network processing~\cite{rad2005real,comaniciu2000real}, 
object and people tracking~\cite{lipton1998moving,stauffer2000learning,karantonis2006implementation},
understanding gestures~\cite{raptis2011real,shotton2013real},
face recognition, speech understanding~\cite{bailenson2008real,crawford2005real},
sound understanding~\cite{saunders1996real,tzanetakis2002musical}
scientific studies~\cite{holmes1984textural,mahabal2008towards}, 
and medical analysis~\cite{rodriguez2005real,laconte2007real}.
All of these applications could benefit from the algorithms 
and indexing structures outlined in this paper.

Our techniques are designed using a wrapper-based approach~\cite{kohavi1997}
that is agnostic to the specific machine learning model, budget metric, and features to extract.
For this reason, our approach can be applied in conjunction with 
a variety of machine learning classification
or regression techniques, including SVMs~\cite{hearst1998support}, 
decision trees~\cite{quinlan1986induction},
linear or ridge regression~\cite{hastie2009elements}, 
among others~\cite{duda2012pattern}.  In addition, the budget can be defined in terms
of systems resources, monetary metrics, time or a combination thereof.

% The key prior work in cost-sensitive real-time classification 
% is a series of papers by Xu et al.~\cite{DBLP:conf/icml/XuWC12,DBLP:conf/icml/XuKHW13,DBLP:conf/icml/XuKWC13,DBLP:conf/kdd/RaykarKY10}. 
% These papers augment several machine learning algorithms
% to include feature extraction costs as a penalty term during the learning process.  This is used
% to define a order of features to extract to meet a cost budget and maximize the expected accuracy.
% The most relevant paper in this collection is GreedyMiser~\cite{DBLP:conf/icml/XuWC12}, which applies this technique to SVM
% models.  Unfortunately, their techniques rely on specially crafting learning models and cannot be used
% for arbitrary models.  In addition, the ordering induced in GreedyMiser does not take into account the fact that real-time costs and input sizes are not fixed,
% thus it produces provably sub-optimal models.
% As we will see in the experiments, this approach indeed yields sub-optimal
% predictive models in practice.

There has been some work on adapting traditional algorithms to incorporate some notion of joint optimization with resource constraints, often motivated by a transition of these algorithms out of the research space and into industry.
A few examples of this approach have been developed by the information retrieval and document ranking communities.
In these papers the setup is typically described as an additive cascade of classifiers intermingled with pruning decisions.
Wang \textit{et al} notes that if these classifiers are independent the constrained selection problem is essentially the knapsack problem.
In practice, members of an ensemble do not independently contribute to ensemble performance, posing a potentially exponential selection problem.
For the ranking domain, Wang \textit{et al} apply an algorithm that attempts to identify and remove redundant features~\cite{ranking-constraints}, assemble a cascade of ranking and pruning functions~\cite{ranking-cascade}, and develop a set of metrics to describe the efficienty-effectiveness tradeoff for these functions~\cite{ranking-efficient}.
Other work focuses specifically on input-sensitive pruning aggresiveness~\cite{retrieval-pruning} and early cascade termination strategies~\cite{early-exit}.
These approaches are similar in spirit to ours but tightly coupled to the IR domain.
For example, redundant feature removal relies on knowledge of shared information between features (e.g., unigrams and bigrams), and the structure of the cascade (cycles of pruning and ranking) is particular to this particular problem.
Further, these approaches are tuned to the ranking application, and do not directly apply to classification.

Xu's classifier cascade work~\cite{classifier-cascade-xu, DBLP:conf/icml/XuKWC13,DBLP:conf/icml/XuKHW13,DBLP:conf/icml/XuWC12} considers the problem of post-processing classifiers for cost sensitivity.
Their approach results in similar benefits to our own (e.g., expensive features may be chosen first if the gains outweigh a combination of cheap features), but it is tailored to binary classification environments with low positive classification rate and does not dynamically factor in runtime input size.
Others apply markov decision processes to navigate the exponential space of feature combinations~\cite{karayev2013dynamic}, terminate feature computation once a test point surpasses a certain similarity to training points~\cite{nan2014fast}, or greedily order feature computation~\cite{naula2014multi}, but none of these formalize the notion of budget or input size into the runtime model, making it difficult to know whether high-cost high-reward features can be justified up front or if they should be forgone for an ensemble of lower-cost features.
That said, our \gm algorithm (along with its variants, \gacc and \gcost) are adapted from these prior papers~\cite{DBLP:conf/icml/XuWC12,naula2014multi}. %\agp{is naula also similar in spirit to \gm? if so, we should mention earlier as well.}

Our \pdom algorithms are also related to prior work on the broad literature on frequent itemset mining~\cite{agrawal1993mining}, specifically~\cite{burdick2001mafia,pasquier1999efficient}, that has a notion of a lattice of sets of items (or {\em market baskets}) that is explored incrementally. Further, portions of the lattice that are dominated are simply not explored.
Our \pdom algorithm is also related to skyline computation~\cite{tan2001efficient}, since we are implicitly maintaining a skyline at all ``interesting'' item sizes where the skyline changes drastically.

\textit{Anytime algorithms} are a concept from planning literature that describe algorithms which always produce \textit{some} answer and continuously refine that answer give more time~\cite{anytime-algorithms}.
Our \gm-family of algorithms are certainly anytime algorithms.
 % Our approach is a slight variant of that setup in that we assume access to a time budget, but this setup could be adapted to the anytime model by initializing a small initial budget and and then iteratively increasing it.

% Argue that
% \squishlist
% \item Standard classification does not work because one classifier orignates as a result of it.

% \item Regularization does not work

% \item latency is an important issue.   the architecture folk work
% on loop perforation (skipping loop iterations to approximate results
% and run faster), implementation alternatives (4 implementations of
% thesame algorithm with different acc/costs).  The IE folk are also
% using different ML solvers for CRF/SVM algorithms that trade off
% acc and cost.
% \squishend

\nocite{*}

\section{Conclusion and Future Work}
In this paper, we designed
machine-learning model-agnostic cost-sensitive prediction schemes. 
We developed two core approaches (coupled with indexing techniques), 
titled \pdom and \gm,
representing two extremes in terms of how this cost-sensitive
prediction wrapper can be architected.
We found that \pdom's optimization schemes allow us to
maintain optimality guarantees while ensuring 
significant performance gains on various parameters
relative to \pdomind, \nlook, and \nexp, and many times
\gm, \gacc, and \gcost as well.
We found that \gm, along with the \gacc and \gcost variants
enable ``quick and dirty'' solutions that are often
close to optimal in many settings.

% titled \pdom, for indexing and retrieving appropriate
% machine learning models for real-time classification.
% The algorithm takes as input the item size and the cost constraints,
% and returns the optimal model given constraints.
% We found that our optimization techniques, both offline and online,
% lead to (a) significant reduction of work (close to XXX in YYY and ZZZ in WWW) in computing the index offline
% (b) a significant reduction of work (close to XXX in YYY and ZZZ in WWW) in retrieving models online, and lastly,
% (c) optimal or near-optimal accuracy of the retrived model (close to XXX as compared to XXX)
% as compared to \gm, \pdomind, \nlook, and \pdomexp.

In our work, we've taken a purely black-box approach towards
how features are extracted,
the machine learning algorithms,
and the structure of the datasets.
In future work, we plan to investigate
how knowledge about the model,
or correlations between features 
can help us avoid expanding even more
nodes in the lattice.

Furthermore, in this paper, we simply used the size of the image
as a signal to indicate how long a feature would take
to get evaluated: we found that this often leads
to estimates with high variance 
(see Figure~\ref{fig:cost_vary_real}),
due to which we had to provision for the worst-case
instead of the average case.
We plan to investigate the use of other ``cheap''
indicators of an item (like the size) 
that allow us to infer how much a feature evaluation
would cost.
Additionally, in this paper, our focus was on
classifying a single point. 
If our goal was to evaluate an entire dataset within a time budget,
to find the item with the highest likelihood
of being in a special class, we would need very different techniques.

% \squishlist 

% \item Dataset structures: in our work we only cared
% about the size of a dataset, however understonding more structured
% information would be desireable.  Suppose each datapoint is a
% database table to be classified in a particular domain.   Sampling
% and computing features over a subset of the records lets the feature extractor
% make a trade off between the cost and accuracy for that particular feature.  
% Understanding how to push down such decisions is interesting.

% \item Model knowledge: if we know how the model works, surely we could do better.

% \squishend

% Balancing columns in a ref list is a bit of a pain because you
% either use a hack like flushend or balance, or manually insert
% a column break.  http://www.tex.ac.uk/cgi-bin/texfaq2html?label=balance
% multicols doesn't work because we're already in two-column mode,
% and flushend isn't awesome, so I choose balance.  See this
% for more info: http://cs.brown.edu/system/software/latex/doc/balance.pdf
%
% Note that in a perfect world balance wants to be in the first
% column of the last page.
%
% If balance doesn't work for you, you can remove that and
% hard-code a column break into the bbl file right before you
% submit:
%
% http://stackoverflow.com/questions/2149854/how-to-manually-equalize-columns-
% in-an-ieee-paper-if-using-bibtex
%
% Or, just remove \balance and give up on balancing the last page.
%
\newpage
{
% If you want to use smaller typesetting for the reference list,
% uncomment the following line:
% \small
%\bibliographystyle{acm-sigchi}
\bibliographystyle{abbrv}
\bibliography{main}
}

\newpage
%!TEX root=../main.tex

\appendix

\section{Proof of Theorem 3.13}

The following a proof of the poly-dom index construction algorithm described in Section~\ref{sec:poly-dom}.

\noindent {\sc Proof}
The proof follows a case analysis:
\begin{compactitem}
\item Scenario 1: Curve 1 and 2 both on skyline and $\alpha_1 > \alpha_2$
\item Scenario 2: Curve 1 is not on the skyline while Curve 2 is
\item Scenario 3: Curve 1 and 2 are both not on skyline 
\item Scenario 4: Curve 1 is on the skyline and Curve 2 is not, and $\alpha_1 > \alpha_2$
\end{compactitem}

The argument is that for Scenario 3 and 4, the skyline will not change:
in scenario 3, Curve 1 and 2 will still not be on the skyline,
while in Scenario 4, since Curve 1 gets even better,
it will still be on the skyline, while Curve 2 will be dominated
by Curve 1 and therefore will not be on the skyline.

On the other hand, for Scenario 1, Curve 1 will start dominating Curve 2,
and so Curve 2 now is removed from the skyline.
For Scenario 2, which is a little tricky, Curve 1, which is not on the skyline
because of high cost, may move into the skyline if there is no other
curve that dominates it (i.e., has lower cost and higher accuracy).

The other scenarios cannot happen:
\begin{compactitem}
\item 
Other half of Scenario 1: Curve 1 and 2 are both on skyline, and $\alpha_1 < \alpha_2$ cannot happen, because then Curve 2
would dominate Curve 1
\item
Other half of Scenario 4:  Curve 1 is on the skyline while Curve 2 is not on the skyline and $\alpha_1 < \alpha_2$ cannot happen
since Curve 1 is dominated by Curve 2 $\square$
\end{compactitem}

\section{Assumption Validation}\label{sec:exp-assumptions}

In addition to the above performance metrics, we evaluated the sources of three types of variation
that deviate from our assumptions.

\begin{figure}[h!]
\vspace{-5pt}
\centering
\subfigure{\includegraphics[width = 1.5in]{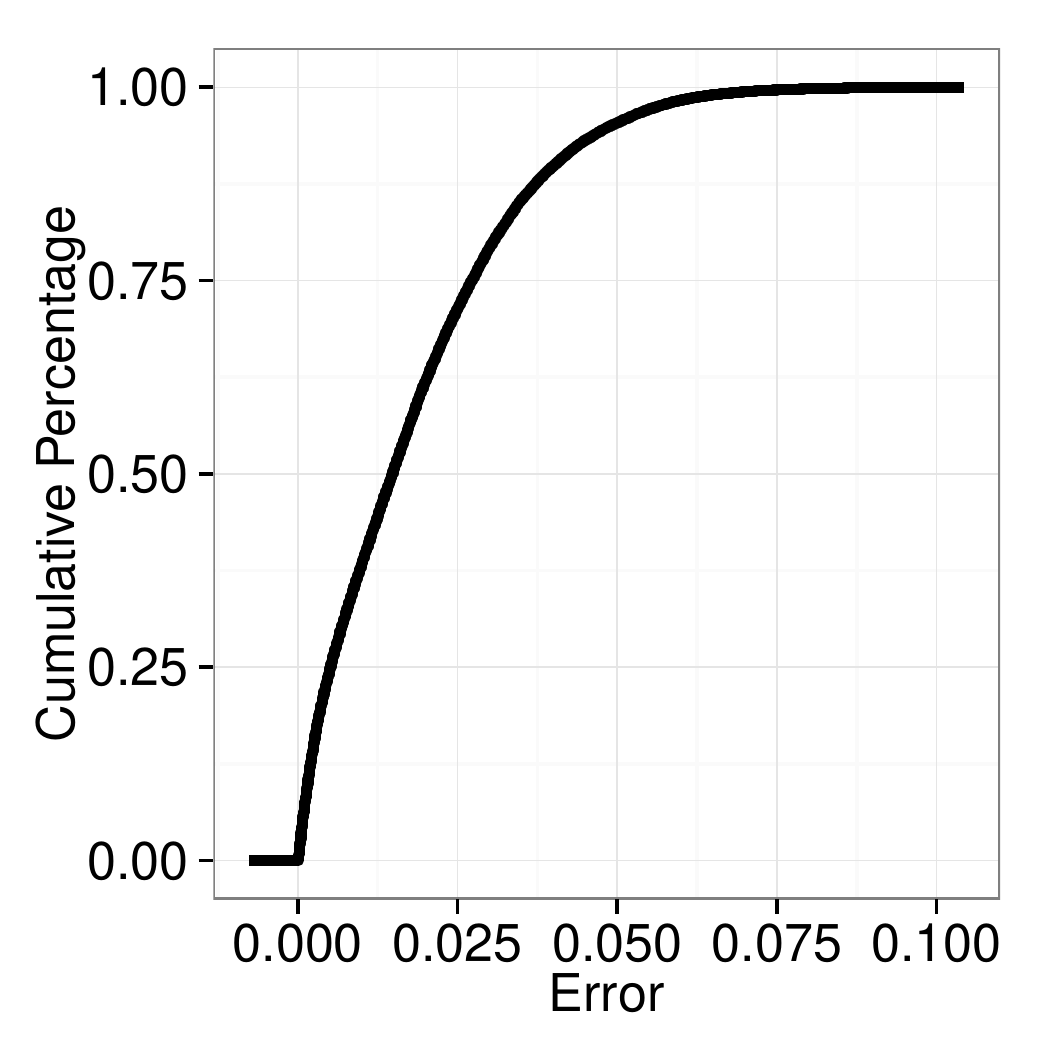}\label{fig:cdf_vary_e_real}}
\subfigure{\includegraphics[width = 1.5in]{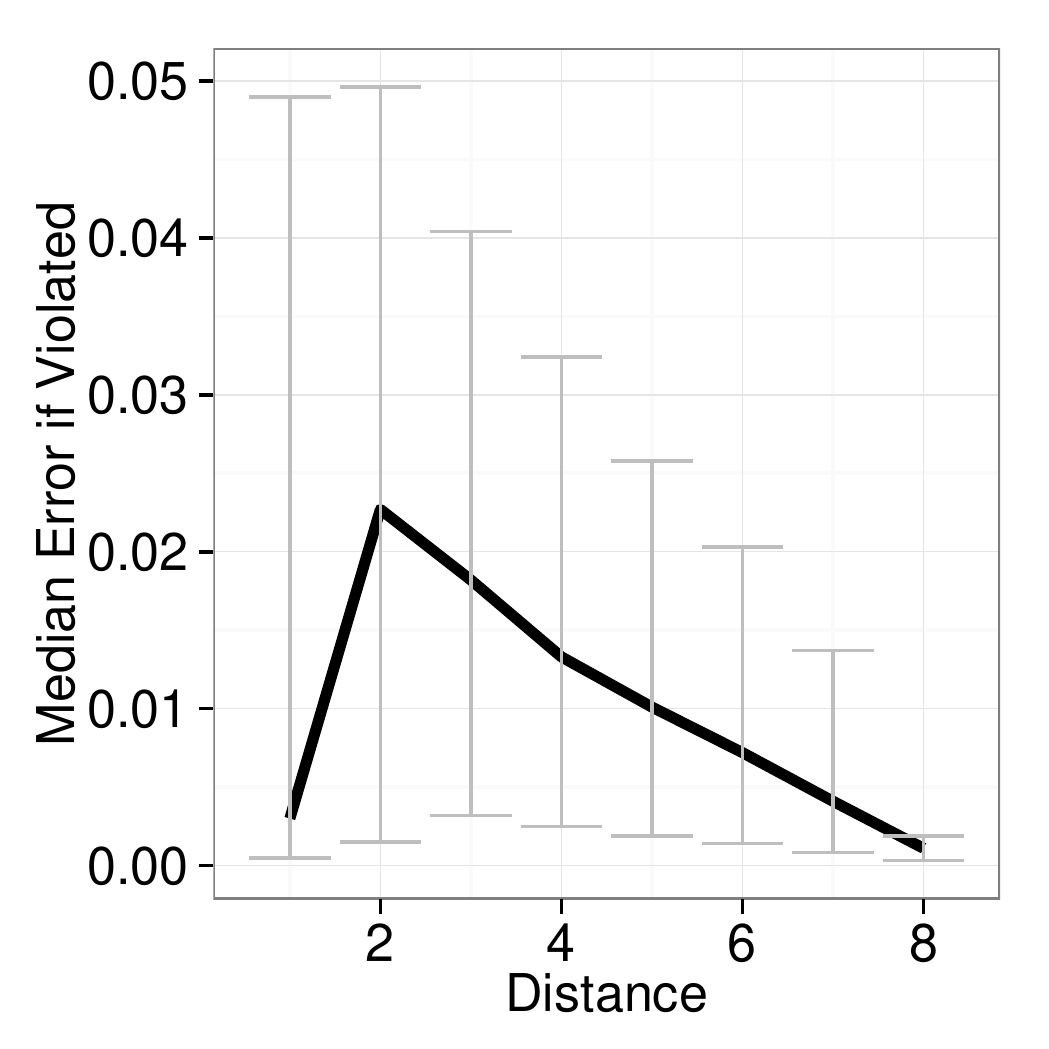}\label{fig:cdf_vary_distance_real}}
%\subfigure{\includegraphics[width = 1.5in]{figures/simulations/percent_vary_distance_real.pdf}\label{fig:percent_vary_distance_real}}
\subfigure{\includegraphics[width = 1.5in]{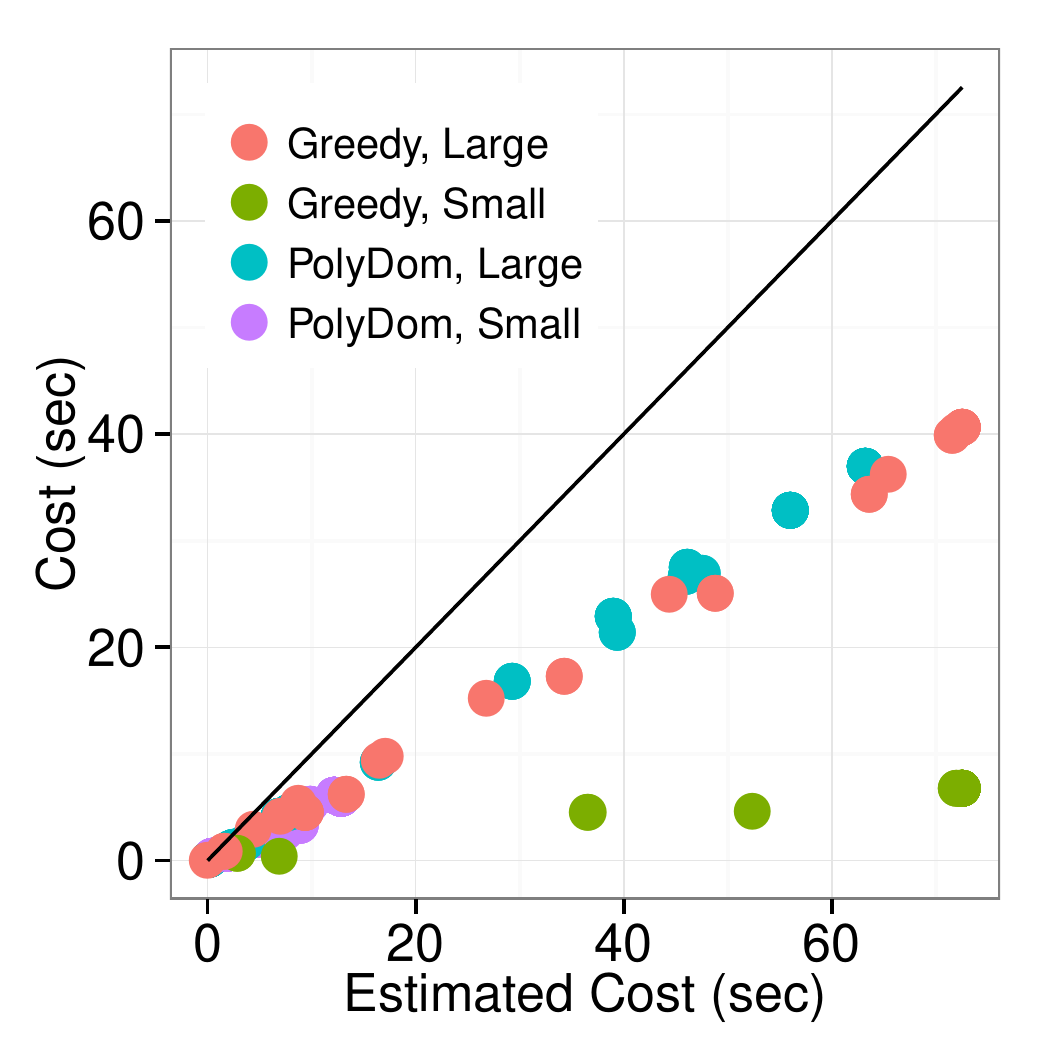}\label{fig:cost_versus_estcost_real}}
\subfigure{\includegraphics[width = 1.5in]{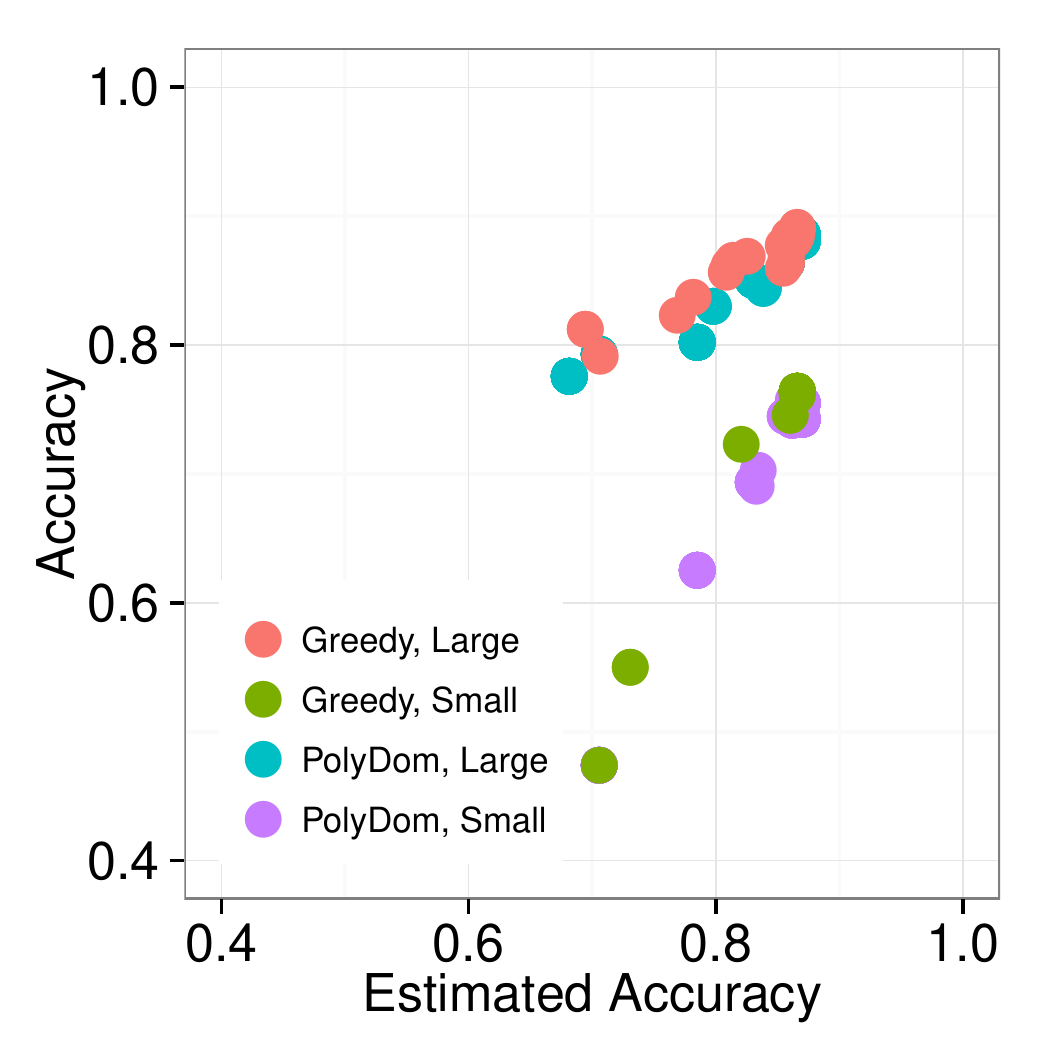}\label{fig:estacc_vs_acc_real}}
\caption{(a) CDF of anti-monotonicity (b) anti-monotonicity as a function of distance 
%(c) Percentage variation of anti-monotonicity with distance 
(c) Cost vs. Estimated Cost (d) Accuracy vs. Estimated Accuracy}
\end{figure}

\stitle{Anti-monotonicity:} First, we focus on the monotonicity
assumption that we made on the real dataset, which we used to define
the $e$ values used in the experiments.  Although the monotonicity
principle is anecdotally known to be true among machine learning
practitioners, we were unable to find a reference for the practical
evaluation of monotonicity on a real dataset.  We view this as an
additional contribution of our paper.

%While we did take some amount of anti-monotonicity
%into account via the $e$ value, we didn't verify whether
%the amount of anti-monotonicity in the real dataset is worse or better than $e$.

Figure~\ref{fig:cdf_vary_e_real}  plots the cumulative distribution
of the violations of monotonicity between pairs of ancestor-descendant feature sets in the lattice.
Most of the violations are small: close to 95\% of the violations
are below an error of $5\%$, while all violations are within $7.5\%$.
Thus, we believe the monotonicity principle is largely true, even on a real dataset.
Note that we found fimor devations between the quality of the retrieved models as we 
decreased $e$ to $0$, so
the assumptions that we made in the paper do not hurt 
us in a significant way.

Next, we would like to evaluate where in fact these violations of monotonicity
exist in the lattice.
Figure~\ref{fig:cdf_vary_distance_real} evaluates the distribution of 
violations as a function of distance\footnote{the difference in number of features between the ancestor and descendant featuresets} between feature set pairs.
The black line is the median $e$ (with grey error bars) of all violations as a function of distance.
As can be seen in the figure, the highest median $e$ as well as the largest variance is at distance $2$
and both quickly decrease to close to $0$ at $8$. 
This is a good sign: violations of monotonicity,
if any, are local rather than global, with almost no violations that
are between pairs of feature sets that are far away from each other.
%Figure~\ref{fig:percent_vary_distance_real} shows similar data where the y axis is
%the percentage of ancestor-descendant feature set pairs with the
%violation of monotonicity of varying levels ($\leq 0.001, 0.01, 0.03, \ldots$).
%We can see that the majority of both large and small errors occur between distances  1---5.
These results suggest that the Skyline algorithm is not likely to falsely prune
a feature set early on due to a violation in monotonicity.  Furthermore, a modest $e$ value
can compensate for the majority of violations.

\stitle{Estimated Versus Actual:} Next, we compare the estimated cost and accuracies of the
real-world model with the true values for large and small image sizes.

Figure~\ref{fig:cost_versus_estcost_real} plots
the estimated versusactual cost.  We find that the cost function tends to 
over estimate the actual cost because the cost functions are trained on the 
worst, rather than average case.  We chose this because if we did provision for the mean cost,
the poly-dom index may return models whose true costs exceed the time budget.
The costs for \gm are similar, however because it ignores item size during
the offline phase, it severely underestimates the cost of the small images, in contrast to \pdom.

Figure~\ref{fig:estacc_vs_acc_real} plots the estimated and true accuracy
of the models retrieved.  We find the the estimated accuracy is indeed linearly correlated with the true accuracy.
However the model consistently overestimates the accuracy because the small images are downsampled, so the features
are correspondingly less accurate. Overall, this suggests that optimizing for estimated accuracy 
is a reliable proxy for the quality of predictions at test time.

\section{Algorithm Pseudocode}

\begin{algorithm}[ht]
 \KwData{${\cal F}, n_0, \alpha$}
 \KwResult{${\sf candidateSet}$}
 ${\sf expandedNodes} =$ {\sc Expand-Enumerate} $({\cal F}, n_0, \alpha)$\;
 ${\sf toRemove} = \emptyset$\;
 \For{s, r $\in {\sf expandedNodes}$}{
	\If{s is dominated by r}{
		add s to ${\sf toRemove}$\;
	} 
 }
 ${\sf candidateSet} = {\sf expandedNodes} - {\sf toRemove}$\;
\caption{{\sc Candidate-Set-Construction}}\label{alg:candidate}
\end{algorithm}

\begin{algorithm}[h!]
  \KwData{${\cal F}, n_0, \alpha$}
  \KwResult{${\sf expandedNodes}$}
  ${\sf activeTop} = \{ {\cal F} \}$\;
  ${\sf activeBottom} =  \{ \{ \} \}$\;
  ${\sf frontierTop} = {\sf frontierBottom} = {\sf expandedNodes}  = \emptyset$\;
  \While{${\sf activeTop}$ or ${\sf activeBottom}$ is non-empty}{
    ${\sf activeTop2} = {\sf activeBottom2} = \emptyset$\;
  	\For{s in ${\sf activeTop}$}{
  		\If{s is not sandwiched between ${\sf frontierTop}$, ${\sf frontierBottom}$ AND has not been expanded}{
  			expand $s$ and add to ${\sf expandedNodes}$\; 
  			add $s$'s children to ${\sf activeTop2}$\;
  			add $s$ to ${\sf frontierTop}$\; remove $s$'s parents from ${\sf frontierTop}$\;
  		}
  		remove $s$ from ${\sf activeTop}$\;
  	}
  	\For{s in ${\sf activeBottom}$}{
  		\If{s is not sandwiched between ${\sf frontierTop}$, ${\sf frontierBottom}$ AND has not been expanded}{
  			expand $s$ and add to ${\sf expandedNodes}$\; 
  			add $s$'s parents to ${\sf activeBottom2}$\;
  			add $s$ to ${\sf frontierBottom}$\; remove $s$'s children from ${\sf frontierBottom}$\;
  		}
  		remove $s$ from ${\sf activeBottom}$\;
  	}
  	${\sf activeTop} = {\sf activeTop2}$\; 
  	${\sf activeBottom} = {\sf activeBottom2}$\;
  }
  \caption{{\sc Expand-Enumerate}}\label{alg:expand-enumerate}
\end{algorithm}

\begin{algorithm}[ht]
  \KwData{${\cal C}, \alpha$}
  \KwResult{${\sf Poly-Dom}$}
  ${\sf candCurves} = $ curves corresponding to ${\cal C}$\;
  ${\sf sortedCurves} = $ sorted ${\sf candCurves}$ for $n < 1$ on cost\;
  ${\sf intPoints} = $ p.queue of int.~pts.~of neighboring curves\;
  \While{${\sf IntPoints}$ is not empty}{
    ${\sf singlePt} = {\sf intPoints}.pop()$\;
    update ${\sf sortedCurves}$ for ${\sf singlePt}$\;   
    ${\sf a, b}$ = curves that intersect at ${\sf singlePt}$\;
    \If{${\sf interesting(singlePt, sortedCurves)}$}{
      ${\sf ptsSoFar}.add({\sf singlePt}$)\;
    }
    ${\sf intPoints}.add($new intersections of ${\sf a, b}$ with neighbors in ${\sf sortedCurves})$\;
  }
  return ${\sf ptsSoFar}$\;
  \caption{{\sc PolyDomIntersections}}\label{alg:poly-dom-int}
\end{algorithm}

\begin{algorithm}[ht]
 \KwData{${\cal F}, n_0$}
 \KwResult{${\sf candidateSeq}$}
 \For{$\lambda \in {\cal L}$}{
  ${\sf curSet} = \emptyset$\;
  ${\sf remSet} = {\cal F}$\;
  \While{${\sf remSet} \neq \emptyset$}{
    add best feature $f \in {\sf remSet}$ to ${\sf curSet}$\;
    ${\sf candidateSeq}(\lambda)$.append(model corresponding to ${\sf curSet}$)\;
    remove $f$ from ${\sf remSet}$\;
  }
 }
 return ${\sf candidateSeq}$\;
\caption{{\sc Greedy Candidate-Sequences}}\label{alg:greedy}
\end{algorithm}

%\newpage

% old content goes here
% \begin{appendix}

% \input{content/xx-problems}
% \input{content/02-approach}
% \input{content/xx-data-structure}
% \end{appendix}

\end{document}